**School of Computing**
FACULTY OF ENGINEERING

UNIVERSITY OF LEEDS

# The performance of multiple language models in identifying offensive language on social media

**Hao Li**

**Submitted in accordance with the requirements for the degree of MSc Advanced Computer Science(Artificial Intelligence)**

**2019/2020**

The candidate confirms that the following have been submitted*:*

| **Items** | **Format** | **Recipient(s) and Date** |
|---|---|---|
| *Deliverables 1* | *Report* | *SSO (27/08/20)* |
| *Deliverable 2* | *Software codes or URL* | *Supervisor, assessor (27/08/20)* |

Type of Project: Theoretical Study / Empirical Investigation

The candidate confirms that the work submitted is their own and the appropriate credit has been given where reference has been made to the work of others.

I understand that failure to attribute material which is obtained from another source may be considered as plagiarism.

(Signature of student)___________________________

# Summary

Text classification is an important topic in the field of natural language processing. It has been preliminarily applied in information retrieval, digital library, automatic abstracting, text filtering, word semantic discrimination and many other fields. The aim of this research is to use a variety of algorithms to test the ability to identify offensive posts and evaluate their performance against a variety of assessment methods.

The motivation for this project is to reduce the harm of these languages to human censors by automating the screening of offending posts. The field is a new one, and despite much interest in the past two years, there has been no focus on the object of the offence. Through the experiment of this project, it should inspire future research on identification methods as well as identification content.

# Acknowledgements

First of all, I would like to thank my parents, Mr. Li Xiaopeng and Ms. Guo Xiaomei. Without their careful guidance in my life, I would hardly be an optimistic and hardworking person. When I feel depressed, they always encourage me. From them, I understand that mastering knowledge is only part of learning, and how to use knowledge to help others is more critical.

Secondly, I wish to express my appreciation to my supervisor, Dr. Brandon Bennett, for his patience and support throughout the process of the project. His words and manner encouraged me. Without his help, the project would not have been completed. When I was confused in my thesis, he always encouraged me and gave me guidance, which enabled me to improve my knowledge of academic writing and machine learning. I also want to thank my assessor, Dr. Marc de Kamps. He carefully read my paper and gave me guidance, and encouraged me to complete the project.

# Table of Contents

## List of Figure

# List of Table

# Chapter 1
# Introduction

## 1.1 Overview

In recent years, as social media has become more and more widely used as a tool for free speech and recording personal life. This is accompanied by the inevitable offensive comments made to users and unabashed hate speech. This article aims to identify whether the tweet is offensive, whether there is a specific object, and what the offending object is by identifying the content of the tweet. Compared with current research in this field, the data set used in this project is more comprehensive, taking into account not only the offending content but also the offending object. Compared with a single algorithm, this project uses a variety of theoretical algorithms including basic classification model, neural network and pre-training model. At the end of the project, a variety of methods including F1-Macro, MCC and obfuscating matrix were used to evaluate various algorithms and demonstrate their performance in this recognition task.

## 1.2 Aim and Objective

The project is to effectively and timely identify offensive comments on social media, such as twitter. By establishing a model to identify whether the tweets on contain offensive language, it can effectively prevent innocent people from being offended by others. For the data mining objectives, identify the offensive word in one sentence. Using classification/prediction algorithms to identify offensive tweets. If it is, categorizing the offending situations and identify the target of the offensive. An OLID data set dedicated to offensive language analysis, which combines the presence or absence of offensive language and potential targets of offensive tasks, will be considered as a good choice for analyzing the performance of various algorithms.

This will involve a number of sub goals such as:

1. Analyze and evaluate the quality of the OLID dataset.
2. Read the literature and materials related to the problem and set a research plan based on it.
3. Data preprocessing is performed on the OLID dataset in order to reduce noise or enhance recognition.
4. Using a small part of dataset to test the effects of data on ALbert models and determine ALbert's advantage. In order to find the most suitable method, multiple techniques will be used for experiments here.

5. Albert, Bi-LSTM and other machine learning and deep learning algorithms for NLP will be use here and I will find better algorithms and parameters through experiments and need to assess the advantages and disadvantages of different techniques.
6. Finally, I will give an evaluation on result and the report to be submit.

### 1.3 Motivation

If visiting the hyde park on a Sunday, we will often see that people standing on folding ladders debating with a crowd of onlookers. Speakers' Corner is a symbol of the UK's centuries-old commitment to free speech. Now, with the development of technology, the Internet has become a new place for personal expression. At the same time, due to the anonymity of social media, many people attack others on the Internet without fear of legal liability. Recent years, hate speech, cyberbullying, or cyberaggression have become pervasive in social media such as Twitter. The use of manual filtering is not only time-consuming but also requires huge funds to pay for expensive workforce and the daily costs of maintaining the company. What is more serious is that it can cause post-traumatic stress disorder-like symptoms to human annotators, so there have been many research efforts aiming at automating the process.

### 1.4 Problem Statement

1. The project has multiple subtasks and aims to identify offensive language on social media from a whole and comprehensive perspective. The first subtask identifies whether twitters is offensive. The second subtask is identified by whether the insulted target has a clear object. The final subtask, an extension of the second subtask, identifies the type of object offended, whether it is an individual, an organization or group.

2. The OLID data set used in this project has 14,200 tweets, but the data is imbalance, and each tweet contains noise that interferes with the identification of offensive language. To solve this problem, we preprocessed the dataset using the Python NLTK package and the open-source code on Github.

3. In this project, data sets are labelled, we will adopt supervised machine learning techniques to identify offensive language.

### 1.5 Deliverables

1. A study of previous research that has been done on the determination, recognition and modeling of offensive language on social media, which are designed to improve the accuracy of recognition.

2. A model for analyzing offensive language based on ALbert and Bi-LSTM algorithm.

3. A report on the model and related techniques experiments.

### 1.6 Project Plan

The report is divided into six chapters and its outline is as follows:

**Chapter 1 :** An introduction of the project with a brief description of the project objectives and the actual timetable.

**Chapter 2 :** Background research includes literature review, technical introduction and technical selection.

**Chapter 3 :** A data understanding that discusses the nature and composition of the data and how to use it to complete testing. The model was designed based on the understanding of the data, building eight different models with the different methods introduced in the background research in order to compare the algorithm's performance.

**Chapter 4 :** Data preprocessing based to the data understanding in the last chapter. Building model based on the algorithm selected in the last chapter, which includes the detailed model structure and parameters.

**Chapter 5 :** Evaluation Experimental on models, using a variety of practical evaluation methods to weigh the performance of the models.

**Chapter 6 :** Project conclusion, discuss its limitations and future improvement, and finally personal reflection.

### 1.7 Project Structure

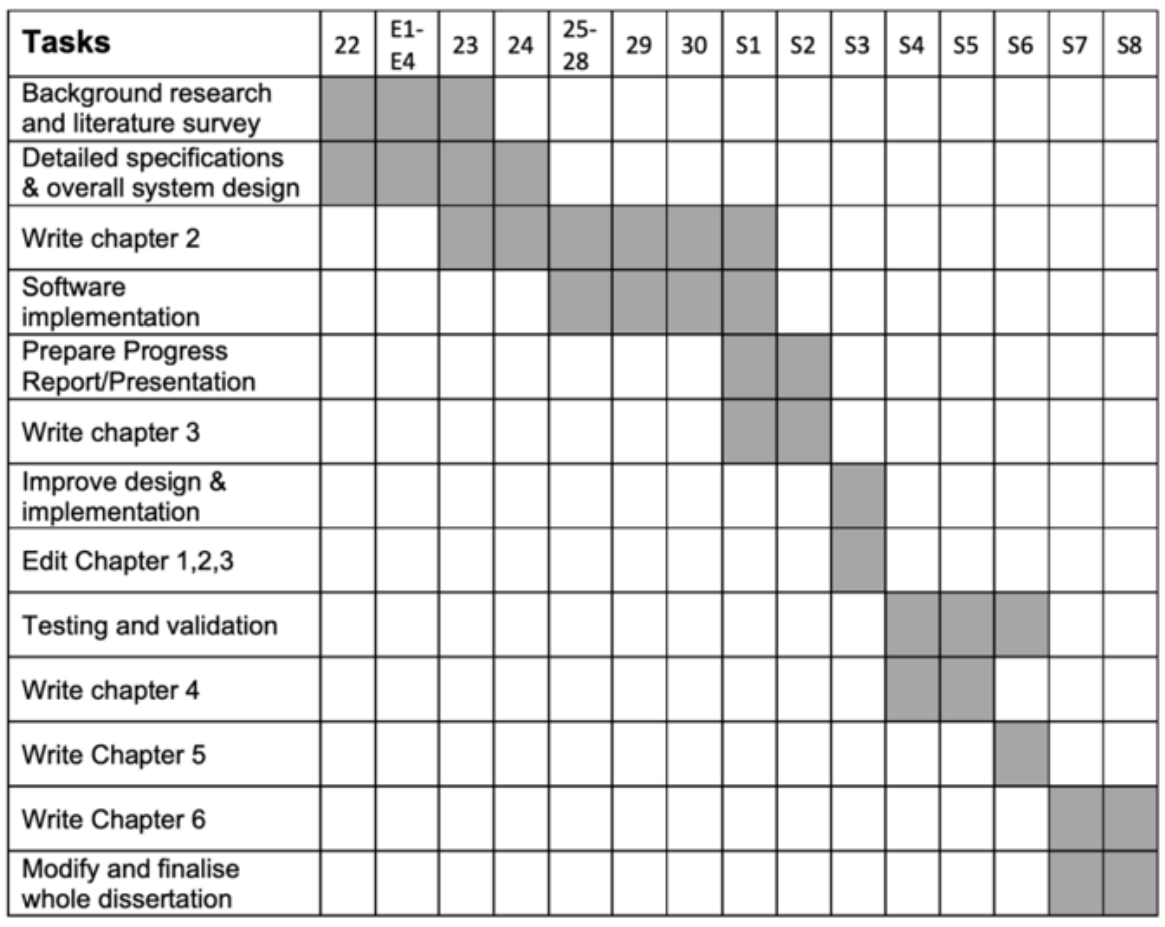

Figure 1.1: Project Structure

# Chapter 2
# Background Research

## 2.1 Summary

This chapter will be divided into three parts. The first section describes the current research and gaps in the field. Research on the use of automated methods to identify cyberbullying has been ongoing for a long time, but there has been no comprehensive study of the various types of offensive language and who they target. In addition, the history of offensive language and the research results of cyberbullying will be introduced in this section. The second section mainly introduces the technical background involved in this project, discusses what machine learning is and the domain background of this project. The last section will discuss the concepts, principles, and reasons for the selection of the algorithms to be used in this project.

## 2.2 Literature Survey

### 2.2.1 Offensive language

Bullying research has been around for a long time, with scientists starting to study bias in work (Morgenstern et al.,1974) and voting (Himmelweit et al .,1978) more than 40 years ago. Since 2000, Bullying has been recognized as a serious problem by the white house (White house,2011). Bullying can include direct or indirect attacks. The former includes physical violence (beating, kicking, using force to obtain items) and verbal violence (mocking, teasing, threatening), and the latter usually includes more subtle, manipulative behavior (such as blackmail, Reject or intimidate others). Another difference between public attacks and relationship attacks is that the former involves cursing, pushing or beating, and the latter involves gossip, spreading rumors, sabotage and other delicate relationships that damage relationships.

The emergence of the Internet allows people to express themselves more freely. However, due to insufficient supervision and the fragmented transmission of emotional extremist information, people tend to make critical remarks based on their subjective judgment when they do not have an accurate and clear understanding of things, resulting in frequent cyber violence.

Cyberbullying refers to the violent behavior carried out by Internet users in the virtual space, which is the extension of violence in the real society. It is a form of violence with serious harm and bad influence. It is a moral judgment that takes place in the virtual space, and it is a general term for the behavior phenomenon of Posting hurtful, insulting and inflammatory words, pictures and videos on the Internet. Compared with the physical violence in people's real life, cyberbullying is a kind of injury and slander to others in the form of words and pictures in the virtual world. What's more, cyberbullying may even have a more profound negative impact on the real life of the person concerned. The participants of cyberbullying are a certain number of netizens, and most of their comments go beyond the normal scope of comments, which are usually accompanied by violations of privacy and crimes.

### 2.2.2 Prior Work

The earliest studies of offensive language began in 2012, when Xu investigated the presence of offensive language on Twitter (Xu et al., 2012). The main contribution of this research is not in the new algorithm, but in demonstrating that social media data and off-the-shelf NLP tools can be an effective combination of bullying research. By breaking down the target into three sub-tasks: whether the post contains offensive language; The role of the person involved in the offending post (is it a victim or a perpetrator); Emotional analysis of offensive posts. Turn offensive language on social media into a familiar NLP task for research. Although the classification model based on SVM and CRF has not achieved good results in this task. F1 scores were 0.36 and 0.47, respectively. The performance was still better than standard off-the-shelf methods, proving that it is feasible to learn from bullying traces. The reason for its poor performance may be that the selected posts is too short and lack of data preprocessing, resulting in a large amount of noise in the data set used.

One of the current strategies for dealing with cyber-attacks is to manually monitor and regulate user-generated content, but the amount and speed of new data on the network makes manual regulation and intervention almost entirely impractical. Thus, Xu's (2012) findings provide the possibility to use automatic methods to identify such behaviors and attract more attention from the research community (Malmasi and Zampieri, 2017).

Nobata et al, (2016) looked at a comprehensive approach to detecting abusive language in user comments. They collected hundreds of comments from Yahoo! Finance and News to create a data set that would show real data. The WWW2015 dataset was also used for comparison with previous work They tried a variety of different features in the task including semantic features and various types of embedding. The results show when these features combined with commonly used NLP features,a desirable results can be produced. Experiments show that the use of Character n-gram alone in these noisy data sets can achieve great results. Several NLP approaches have been used in previous work, but these characteristics have never been combined or evaluated mutually. They took an important step forward in this area by first providing a well-planned public data set and also performing a series of assessments of NLP characteristics.

Kumar et al, (2018) research focused on how to identify cyber-aggressive. Using 12, 000 random comments from Facebook as a data set and an additional 3, 000 comments in English and Hindi as test sets, they tried to sort Aggressive into three classes: Overtly Aggressive (OAG), Covertly Aggressive (CAG) and Non- Aggressive (NAG). The performance of the best systems in their research show that aggression identification is difficult task to solve. In addition , the performance of neural network-based systems do not seem to differ in the performance of other approaches. If features are carefully selected, classifiers such as support vector machines, and even random forests and logistic regression perform as well as deep neural networks.

Dadvar et al., (2013) analyzed a case of cyberbullying in a YouTube comment. Based on content, specific network attacks, and user-specific characteristics, they used SVM to classify whether cyberbullying existed or not. The results show that adding context in the form of a user's activity history can improve the accuracy of cyber-bullying detection.

Djuric et al., (2015) propose a two-step method for hate speech detection. Malmasi and Zampier (2017) uses three different features in the model to identify hate speech.

There has been a great deal of interest in the field from different angles, resulting in a comprehensive understanding and terminology of the issue. On the one hand, it provides us with a very rich and broad view of the phenomenon. On the other hand, it creates a theoretical gap in understanding the interrelationships between these phenomena. In addition, it leads to a degree of duplication of studies, with different studies focusing on the same problem lead to lacking characteristics. In order to optimize solutions to such complex problems, at least to avoid duplication of useless research, standardized data sets and evaluation of different methods are needed.

Recently, Waseem et al. (2017) analyzed the similarities between different approaches proposed in previous work and argued that an model should be designed to distinguish whether offensive language refers to a particular person or group, and whether there is explicit or implicit content.. Wiegand et al. (2018) further applied this idea to German tweets. Their research focused on testing whether tweets are offensive. Further, a second task was designed to categorize offensive tweets as blasphemous, abusive or insulting. However, to the best of our knowledge, no research has explored the targeting of offensive language, which may be important in many cases, such as when studying hate speech aimed at characteristic targets. This project will bridge this gap and test the performance of multiple models in this task.

## 2.3 Methods and Technology

### 2.3.1 Machine Learning

Machine learning is a branch of artificial intelligence that focuses on enabling machines to learn patterns from past experiences, model them based on data, and make predictions about the future.(Alpaydin et al., 2020). Generally, Machine learning has a wide range of uses, including categorization tasks for predicting categories, regression tasks for predicting values based on samples, and reinforcement learning for algorithms to learn by themselves based on results.

### 2.3.2 Deep Learning

Deep learning is a new subfield of machine learning, inspired by the brain structure and function of artificial neural networks. Unlike the neural networks of the 1990s, deep learning base on a huge scale which often with huge neural networks and massive amounts of training data.

Except its scale, another advantage of deep learning is feature learning. They can perform automatic feature extraction from raw data. It tries to use the unknown structure in the input to find the good performance features, and the higher features are represented by the lower features. Hierarchical structure plays an important role in feature learning . It is usually self-directed learning at multiple levels, which combines low-level features into higher-level features for learning.

### 2.3.3 NLP

Natural language processing is an important direction in the field of computer science and artificial intelligence. It research how to effectively use natural language to communicate between humans and computers. Text classification is the most common and important task type in the field of NLP applications. Its purpose is to classify the texts according to certain rules. The "rules" can be determined by people, or algorithms can be automatically summarized from labeled data. Generally speaking, we will first consider artificial regulations, and if it is not feasible, we will consider using algorithms. In our lives, many things can be transformed into a classification problem to solve, such as "is this solution good" can be transformed into a two-classification problem. The same is true in the field of natural language processing. A large number of tasks can be solved by text classification, such as spam text recognition, violent text recognition, text matching, named entity recognition, etc.

### 2.3.4 Supervised learning & Unsupervised learning

Supervised learning is a common method in machine learning tasks. It learns predictive patterns from label training data and builds models. The label refers to the input and output are determined and get the desired output through training. Unsupervised learning, by contrast, extrapolates results from unlabel training data. The most typical unsupervised learning is cluster analysis, which explores hidden patterns. Figure 2.1 shows the relationship between them, supervised learning when the data is labeled. Otherwise, it is unsupervised learning.

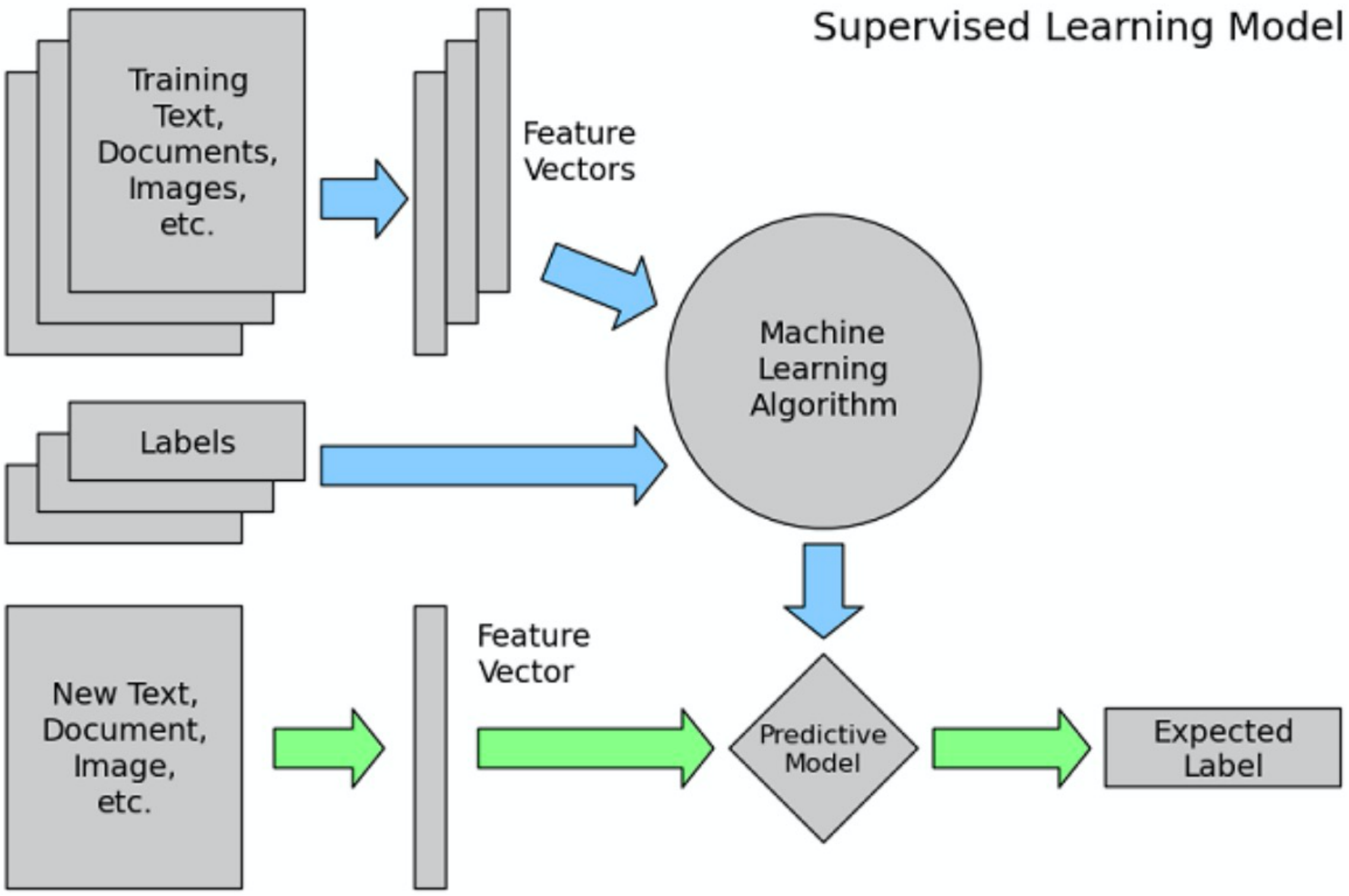

Figure 2.1 Supervised learning model (Kumar ,2018)

### 2.3.5 One-Hot Encoding

One Hot encoding is a form of converting category variables into machine learning algorithms that are easy to use. It mainly uses N-bit state registers to encode N states. Each state has its own register bit, and only one bit is valid at any time. In other words, there's only one place that's 1 at a time, and everything else is 0.

In actual machine learning tasks, features are sometimes not always continuous, but may be categorical values, such as gender. In machine learning tasks, we usually need to digitize such features, as shown in the following examples:

Gender: [" Male ", "Female "]

Area: [" Europe ", "US", "Asia"]

For a sample, such as ["male", "US "]. The gender attribute is two-dimensional and the regional attribute is three-dimensional. We use one-hot encoding, and the feature becomes' male '[1,0]; 'Area' [0,1,0]. The complete feature expression is [1,0,0,1,0]. In this way, all features have a unique expression.

### 2.3.6 N-gram

N-gram is an algorithm based on statistical language model. Usually, We think the current word in a sentence as related only to the n-1 word that precedes it. So it's common to use a N length gram gliding through the sentence to get a sequence of length N by. In the text, the occurrence probability of sentence S is composed of the occurrence probability of N items of S. Through a preset threshold, the frequency of all n-gram occurrences is counted to obtain a list of gram which is the vector eigenspace of the text. There are three main types of N-gram commonly used: the unigram, the Bigram, and the trigram that is, the gram is one, two, three.

If we have a sequence of M words (or a sentence), we want to calculate the probability, according to the chain rule:

$$p(w_1, w_2, \dots, w_m) = p(w_1) * p(w_2|w_1) * p(w_3|w_1, w_2) \dots p(w_m|w_1, w_2, \dots, w_{m-1}). \qquad (2.1)$$

This probability is not easy to calculate, so we use the Markov chain assumption that the current word is only related to a finite number of previous words. Therefore, it is not necessary to go back to the original word, and greatly reduce the length of the above equation.

$$p(w_1, w_2, \dots, w_m) = p(w_i|w_{i-n+1}, \dots, w_{i-1}). \qquad (2.2)$$

In this way, the calculation amount and time in the medium are balanced.

### 2.3.7 Weka

WEKA is an open source machine learning and data mining software based on Java. The software integrates a large number of statistical machine learning algorithms, and users only need to adjust the parameters to use it quickly. Versatility is one of Weka's strengths, almost tasks in machine learning can be found solution in the software.

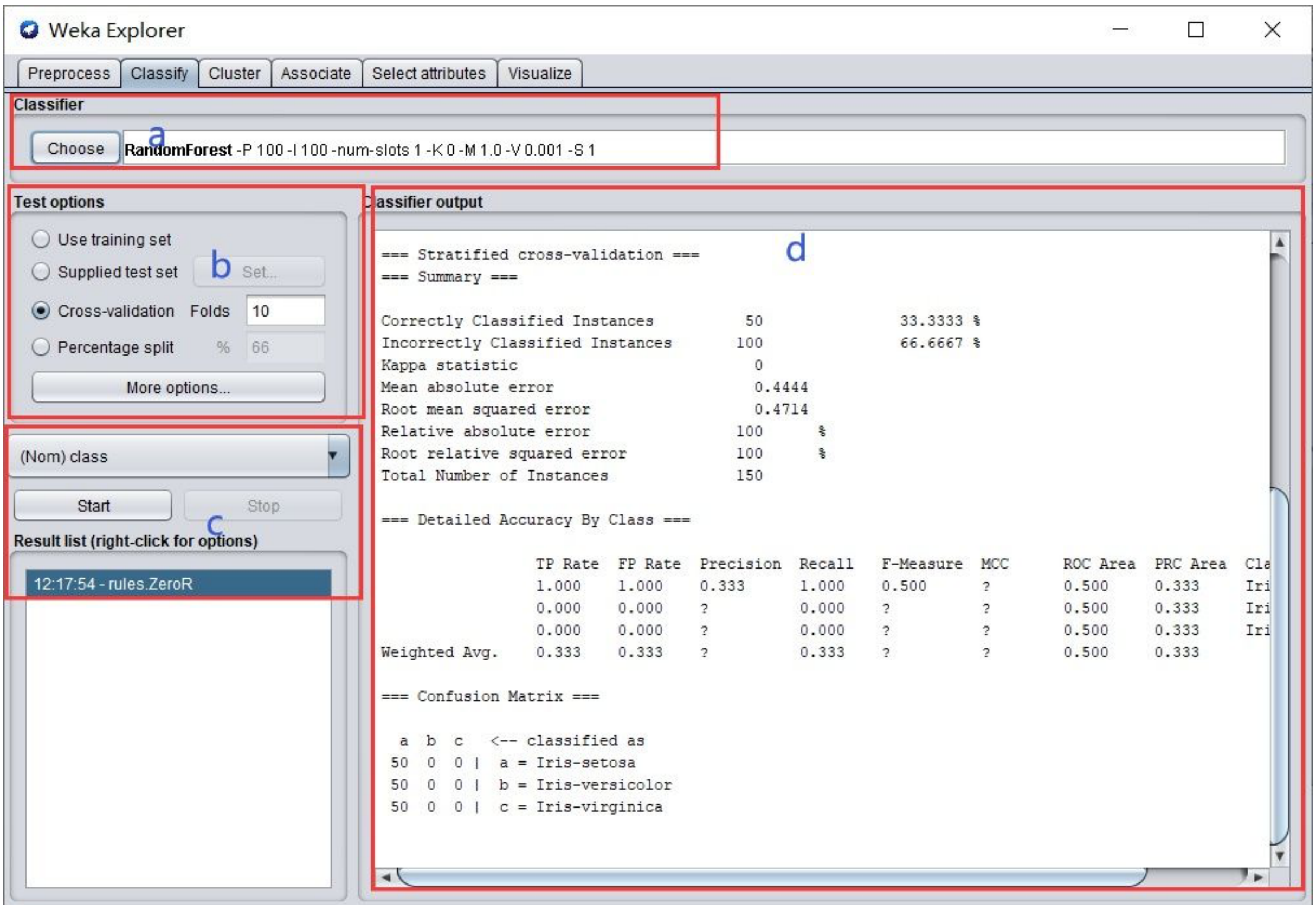

Figure 2.2: The interface of Weka

It is easy to use, complete the machine learning process by operating on the user interface and visually present the results. Take Random Forest algorithm as an example. In Figure. 2.16, A is the classification algorithm to be used. Double-click on the blank of A to pop out the parameters of the corresponding algorithm and adjust the parameters by yourself. B is to set some parameters of model training, such as cross-validation or proportional validation; C starts running the model; D precision statistics of operation results.

### 2.3.8 Word2Vec

In natural language processing, the language model cannot directly process words, so words need to be converted into word vectors for calculation. Common embedding methods include one-hot embedding, etc. Although one-hot method is simple and easy to process text, its vectors are too sparse and waste a lot of vector space. So, I chose to use Word2vec, which maps text to a smaller vector space and keeps each word unique, as a word embedding tool. Word2vec is an efficient tool for characterizing words as real-valued vectors. There are two models used: CBOW and Skip-Gram (Rong, 2014). Through training, word2Vec will map text to n-dimensional vector space, where the similarity in vector space can be used to represent the semantic similarity of text. Based on this principle, word2vec has important applications in many NLP related tasks. (Mikolov et al., 2013).

#### 2.3.8.1 CBOW

Continuous Bag-of-Words Model is a model similar to forward Neural Network Language Model(Bengio, 2003). The difference is that CBOW removes the most time-consuming nonlinear hidden layer and all words share the hidden layer. The figure below shows the CBOW model of multiple context words. When calculating the output of the hidden layer, CBOW does not directly copy the input vector of the context, but averages the input vector of the context word and then uses it as the output.

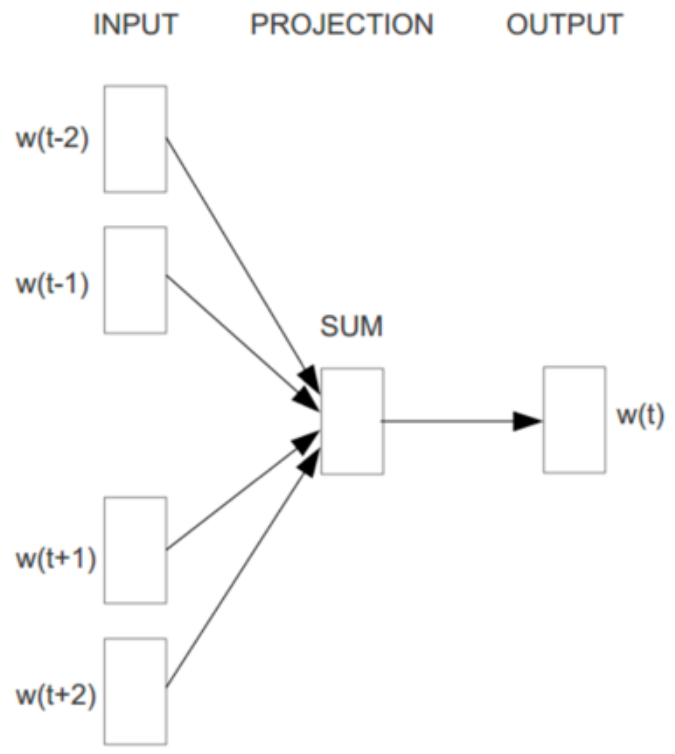

Figure 2.3 The process of CBOW (Rong, 2014)

#### 2.3.8.2 Skip-gram

The Skip-gram model can be regarded as the inverse process of the CBOW model. The target word of the CBOW model is used as input, and the context is used as output.

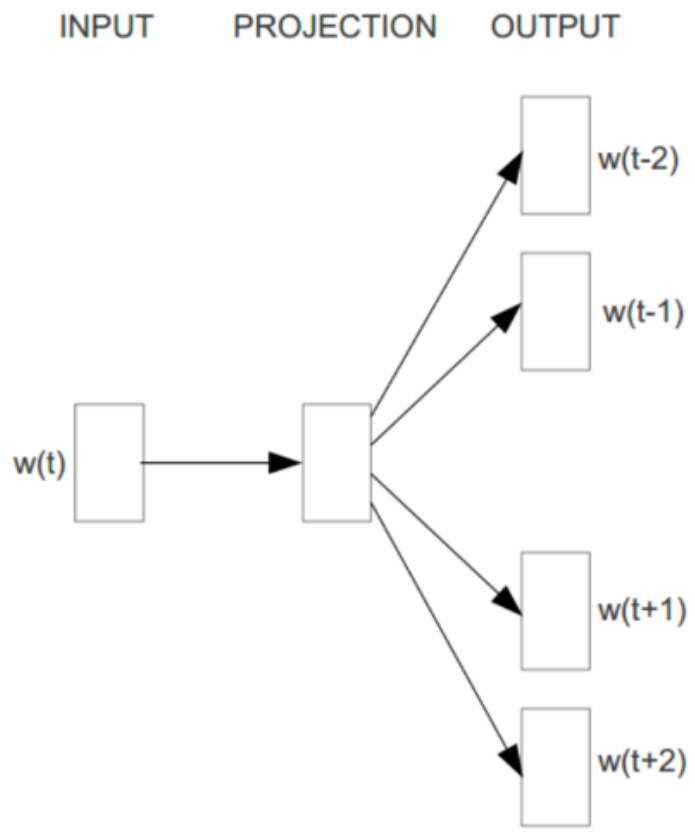

Figure 2.4 The process of Skip-gram (Rong, 2014)

### 2.3.9 Pre-train model

Pretraining models have performed well in natural language processing tasks in the past two years. Previous researchers have designed a benchmark model for us. We can easily use this pre-trained model on our own NLP data set instead of building a model from scratch to solve similar NLP problems. In addition, using these most advanced pre-trained models can help us save a lot of time and computing resources. In order to test the performance of models based on different theories in identifying offensive language, the latest pretraining models ALbert were considered to be included in the experiment.

### 2.3.10 Transformer

BERT algorithm used in this project to generate word vectors has achieved significant improvement in 11 tasks of NLP, which can be called the most exciting findings in the field of deep learning in 2018. The most important part of BERT's algorithm is the application of the concept of Transformer. Attention mechanism was proposed by Vaswani et al.(2014) and has been widely applied in various fields of deep learning in recent years. Transformer is the most important application. Traditional CNN and RNN have been abandoned in Transformer, and the entire network structure is composed entirely of the Attention mechanism. To be more precise, Transformer consists of self-attenion and Feed Forward Neural networks only The author's experiment is to build a total of 12 layers of Encoder-Decoder, and achieve a new high BLEU in machine translation. The reason for the adoption of the Attention mechanism is that RNN (or LSTM, GRU, etc.) can only be calculated sequentially. More specifically, RNN can only calculate from one direction.This mechanism brings two problems:

1.The calculation of time slice $t$ depends on the calculation result of time $t-1$, which limits the parallel ability of the model.

2.In the process of sequential computation, information will be lost. Although the structure of LSTM gate mechanism alleviates the problem of long-term dependence to a certain extent, LSTM is still powerless for the phenomenon of especially long-term dependence.

Transformer solves both of these problems by using the Attention mechanism to reduce the distance between any two positions in a sequence to a constant. Secondly, it is not a sequential structure similar to RNN, so it has better parallelism and conforms to the existing GPU framework.

### 2.3.11 Naive Bayes

Naive Bayes is the simplest form of Bayesian network, which is a supervised learning algorithm to solve classification problems (Harry Zhang). It is often used in classification tasks, so we added it to the experiment to analyze its performance in this project. The advantages of this algorithm are its simplicity, high learning efficiency, and the effect of classification in some fields can be the same as that of decision tree and neural network. On the other hand, the accuracy of the algorithm is affected because the algorithm assumes the independence between independent variables and the normality of continuous variables.

$$P(Y|X) = \frac{P(X|Y)P(Y)}{P(X)} \tag{2.3}$$

X is attribute set, Y is class variable, P(Y) is prior probability, P(X|Y) is conditional probability, P(X) is evidence, P(Y b| X) is posterior probability. Bayesian classification model is to use prior probability P(Y), quasi-conditional probability P(X|Y) and P(X) to represent the posterior probability.

Since the sample in the dataset used is posts rather than images, we used discrete naive Bayes. In other words. The prior probability of each class is estimated using the occurrence frequency of the sample in the training set.

MultinomialNB assumptions prior probability for polynomial distribution characteristics, namely the following type:

$$P(X_j = x_{jl}|Y = C_k) = \frac{x_{jl}+\lambda}{m_k+n\lambda} \tag{2.4}$$

Among them, the $P(X_j = x_{jl}|Y = C_k)$ is the first k category of the $j$ d characteristics of the $l$ all conditional probability values. $m_k$ is central output for the first k class training sample number. $\lambda$ as a constant greater than zero, often take 1, namely, Laplacian smoothing. We could have taken other values.

### 2.3.12 K-NearestNeighbor

K-nearest neighbour classification is one of the clustering algorithms. Since there is no need to obtain prior probability, it is very suitable for the selection of the classification task. In this project, the labeled data are all helpful for KNN to form clustering, so we add this algorithm into the experiment to observe its performance.The main idea of KNN is very simple: for a new sample, if the K samples closest to it belong to a certain category, it is considered that the sample also belongs to that category.

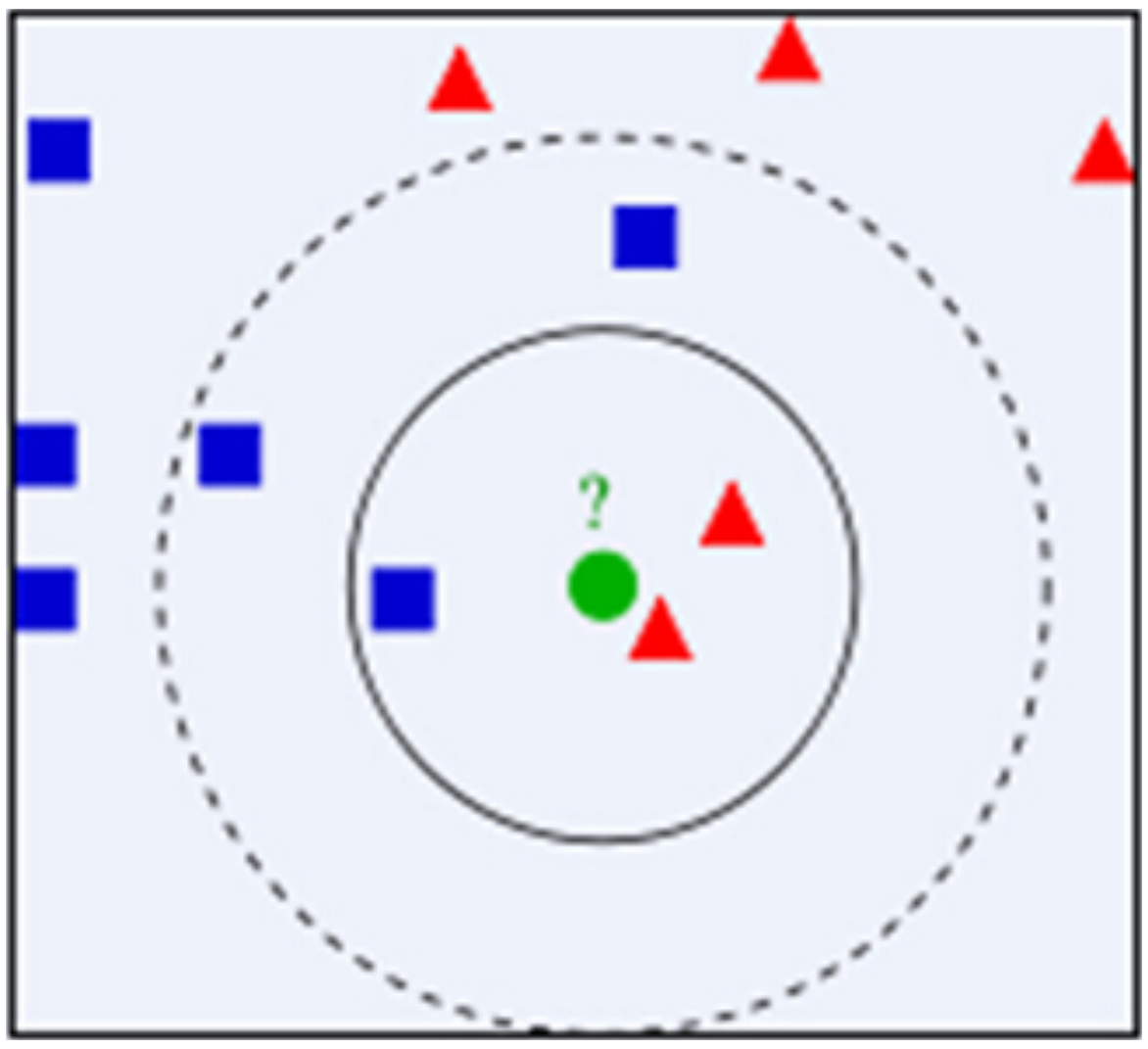

Figure 2.5 The example of KNN

In KNN, the matching between objects is avoided by calculating the distance between objects as the non-similarity index between objects. Generally, the distance is euclidian distance or Manhattan distance:

$$\text{Euclidian distance}: d(x,y) = \sqrt{\sum_{k=1}^{n}(x_k - y_k)^2} \quad (2.5)$$

$$\text{Manhattan distance}: d(x,y) = \sqrt{\sum_{k=1}^{n}|x_k - y_k|} \quad (2.6)$$

KNN algorithm ideas: Using labeled data, the input data will be compared according to the characteristics of the data in the training set to determine the most similar multiple data, and the final classification will be the class with the most frequent occurrence of the data. The specific description of the algorithm is as follows:

1) Calculate the distance between test data and training data;

2) Data are arranged in increasing order according to the distance between them.;

3) Choose K points with the smallest value according to the distance.;

4) Calculate the occurrence frequency of the category of the first K points

5) The final predicted category is the most frequent among the first K points.

It can be seen that the result of KNN algorithm depends on the choice of K

**2.3.13 SVM**

Support vector machine (SVM) is a classic tool for solving machine learning tasks by means of optimization methods. Since the first two subtasks in this task are binary classify tasks, SVM, which is specially designed for binary classify task, can be used as the base line to measure the advantages and disadvantages of the pre-training model we mainly researched. Originally proposed by Cortes et al.(1995) SVM has made great progress in theoretical research and algorithm implementation in recent years,and is beginning to become a powerful means to overcome the "dimensional disaster" and difficulties of over-learning. The SVM used for classification is essentially a binary classification model. SVM belongs to supervised learning. The purpose of SVM is to find a hyperplane in a given sample set containing positive and negative examples to segment the positive and negative examples in the sample, while ensuring the maximum interval between the positive and negative examples. In this way, the classification results are more reliable and the classification prediction ability is better for unknown new samples. As can be seen from the Figure 2.6, in order to maximize the distance between the classes, we don't need to consider all the points, we just need to make the points closest to the separated hyperplane as far as possible from the separated hyperplane. It makes sense to think that the points closest to the separated hyperplane are support vectors.

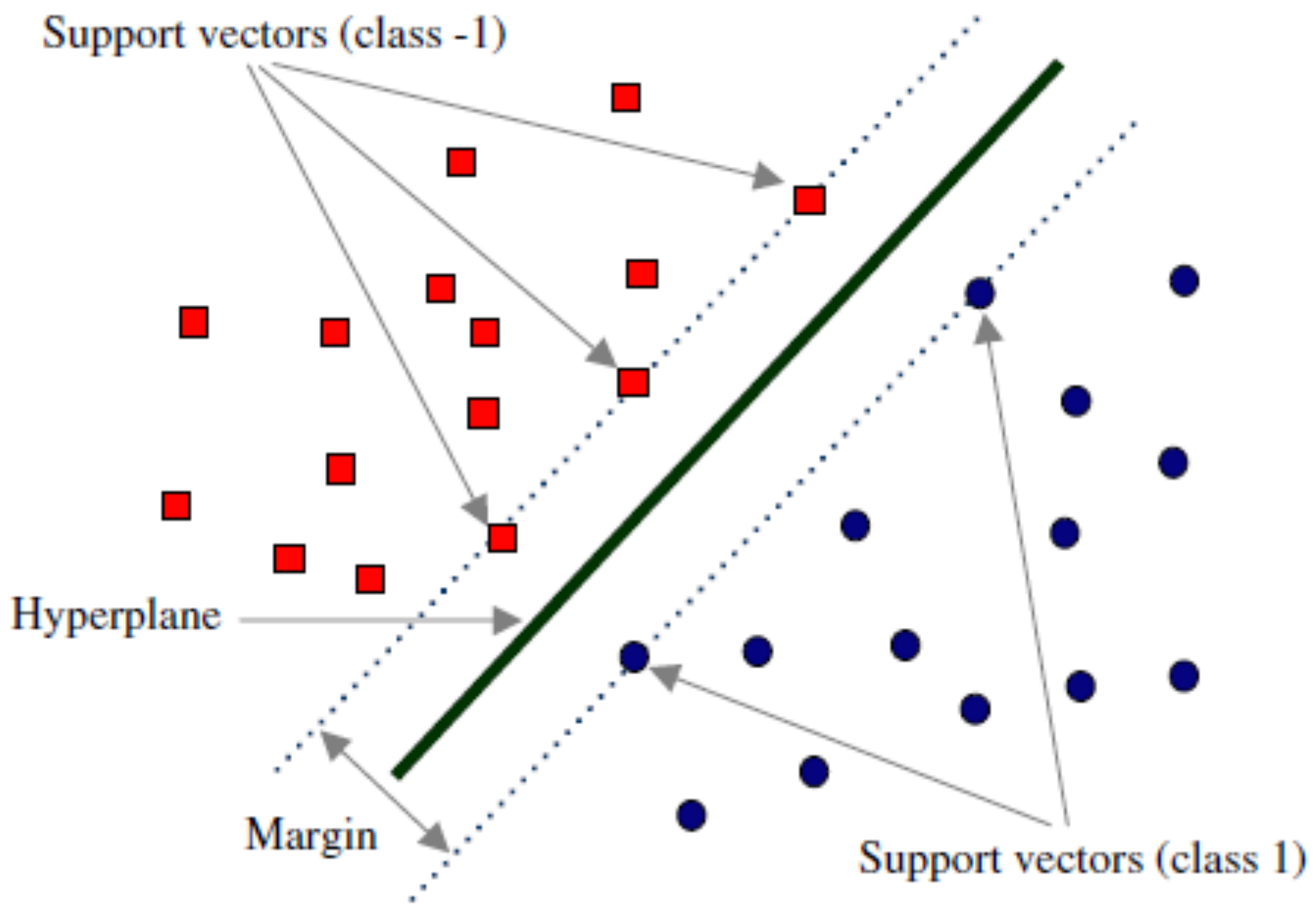

Figure 2.6 The example of SVM

The Support Vector Classification is a model specially used for Classification tasks. In this project, SVC is mainly considered for the identification of offensive language.

**2.3.14 Decision Tree**

Decision trees, like KNN, are algorithms that deal with classification problems. Compared with KNN, decision trees perform better when dealing with unbalanced data. It should be noted that after analyzing the data set, we found that the data for all three tasks were unbalanced. Therefore, the decision tree is considered in this project. It represents a mapping relationship between object attributes and object values, and we can use the decision tree to discover the knowledge contained in the data.

The first step in constructing a decision tree is to select features to divide the original data into several data sets, so each feature must be evaluated. And then through the characteristics of the original data is divided into several data subset of the data subset distribution in all branches of the first decision point, if all the data for the same type on the branch, is divided to stop, if all the data on the branch is not the same type, then also need to continue, until all have the same type of data in a data subset. When dividing with decision tree, the key is which feature is selected to divide each time. When dividing data, we must use quantitative method to judge how to divide data.

The key of decision tree learning lies in how to choose the optimal partition attribute. For binary classification, it is to make the divided samples belong to the same category as far as possible. To measure the quality of features, "information entropy" and "information gain" are used.

Information entropy is used to measure the amount of information in an attribute.

$$Entropy = Entropy(p_1, p_2, p_3 \dots, p_n) = -\sum_{i=1}^{n} p_i \log_2 p_i \quad (2.7)$$

The smaller the entropy represent the better the distribution of the sample to the target attribute. On the contrary, the higher the entropy is, the more chaotic the target attribute distribution of the sample is.

The information gain is the difference between the entropy of the sample data set before partition and the entropy of the sample data set after partition.

$$Gain(S, A) = Entropy(S) - Entropy_A(S) \quad (2.8)$$

Decision tree construction algorithms mainly include ID3, C4.5 and CART, among which ID3 and C4.5 are classification trees and CART is classification regression tree. ID3 is the most basic decision tree construction algorithm, while C4.5 and CART are optimization algorithms based on ID3.

### 2.3.15 Random Forest

The random forest is similar to the decision tree, which is composed of multiple decision trees. Compared with decision trees, the advantage of random forest is that the results are more reliable. Each input data throughout each decision tree and returns a separate classified result. The random forest will take the class with the most classification from decision tree as the final result.

### 2.3.16 Logistic Regression

Logistic regression is usually used to solve regression tasks. Because of simplicity, interpretability, it is often used to classify tasks. The idea of logistic regression is to assume that the data obey Bernoulli's distribution and introduce nonlinear factors, so that dichotomy problems can be dealt with easily.

$$\text{Sogmoid function } g(z) = \frac{1}{1+e^{-z}} \quad (2.9)$$

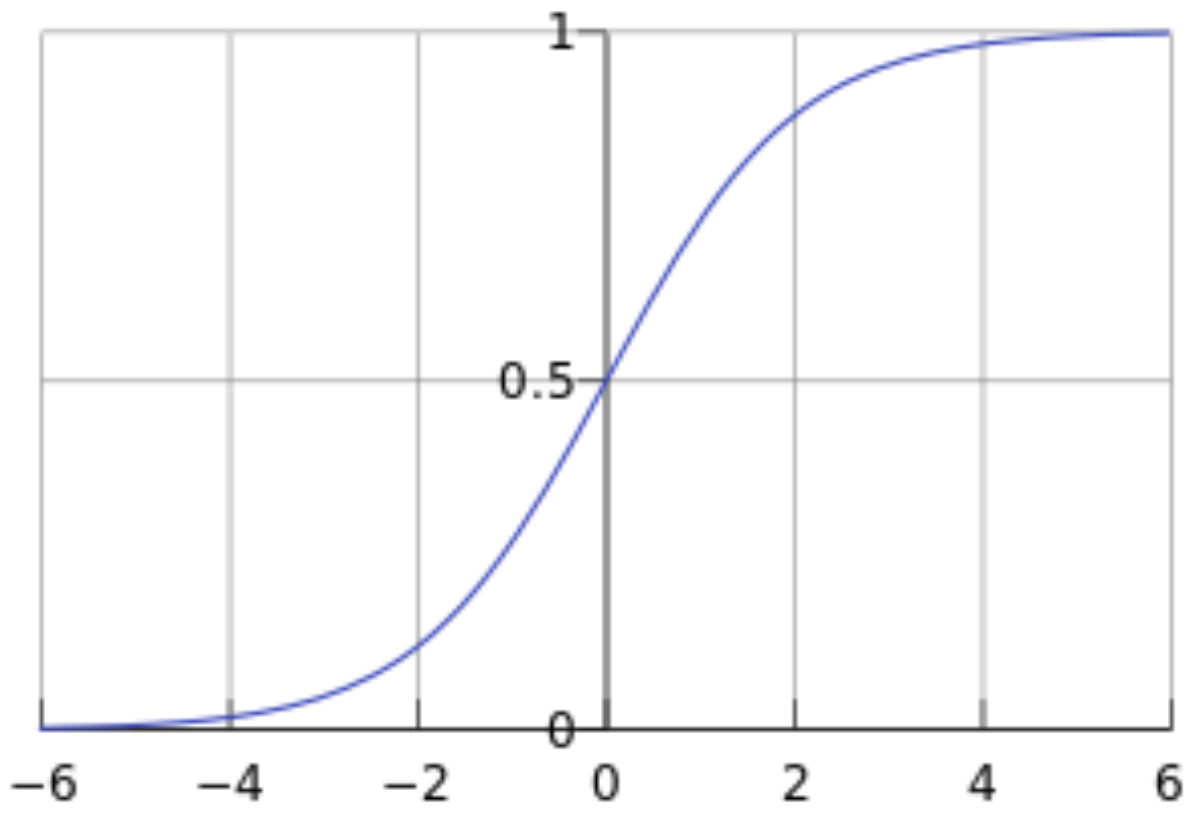

Figure 2.7: The sigmoid function curve

As shown in the Figure 2.7, the sigmoid function is an S-shaped curve, and its value is between [0, 1]. At the distance from 0, the value of the function will quickly approach 0 or 1. This feature is important to solve the dichotomy problem.

The hypothetical function form of logistic regression is as follows:

$$h_\theta(x) = g(\theta^T x) = \frac{1}{1+e^{-\theta^T x}} \quad (2.10)$$

Where $x$ is our input, $\theta$ is the parameter we want to take.

### 2.3.18 Bi-LSTM

Compared with the previous algorithm, LSTM (Long Short Term Memory Networks) is a special kind of RNN (Recurrent Neural Network). Because of its design features, LSTM is suitable for modelling temporal data, such as text data. However, there is a problem with modeling sentences using the LSTM: the model cannot encode information from back to front. In the more fine-grained classification, such as the five classification tasks of strong positive sense, weak positive sense, neutral sense, weak negative sense and strong negative sense, attention should be paid to the interaction among emotion words, degree words and negative words. For example, "This restaurant is big enough" The "enough" here is a modification of the level of “big.” Bi-LSTM (Kang, 2017) can better capture two-way semantic dependency. As the best performing algorithm in the previous generation of language models, it is considered to conduct experiments to compare the performance of various algorithms.

LSTM model is composed of input $x_t$ at the time t, cell state $c_t$, temporary cell state $c_{tt}$, hidden layer state $h_t$, forget gate layer $f_t$, memory gate $i_t$ and output gate $o_t$. LSTM calculation process can be summarized as: based on the cell state to forget old information and remember new information, gives useful information for subsequent moment calculation. LSTM output state of the hidden layer in every time step and discard useless information. Forgetting, memory and output are controlled by forgetting gate $f_t$, memory gate $i_t$ and output gate $o_t$.

which are calculated from the hidden layer state $h_t$at the last moment and the current input $x_t$ (Kang, 2017).

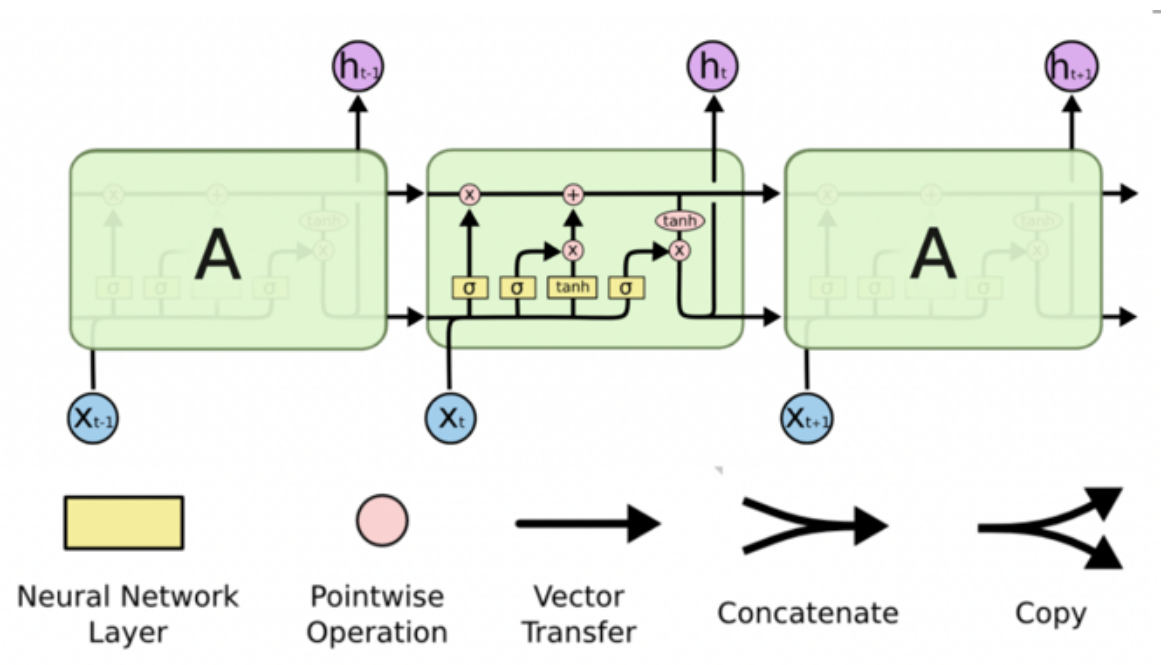

Figure 2.8: The Main Body in LSTM structure (Kang, 2017)

Bi-LSTM is formed by combining the forward LSTM and the backward LSTM. For example, we encode the phrase "I love You", as shown in figure. Forward LSTML input "I", "love" and "You" in turn to get three vectors $\{h_{l0},\ h_{l1},\ \ h_{l2}\}$. Then enter "You", "love" and "I" into the LSTMR, and we will get three vectors $\{h_{r0}, h_{r1}\ , h_{r2}\ \}$. Finally, the forward and backward implicit vectors are spliced together to get {[$h_{l0}, h_{r2}$], [$h_{l1},\ \ , h_{r1}$], [$h_{l2}, h_{r0}$]}, that is, $\{h_0\ ,h_1, h_2\}$.

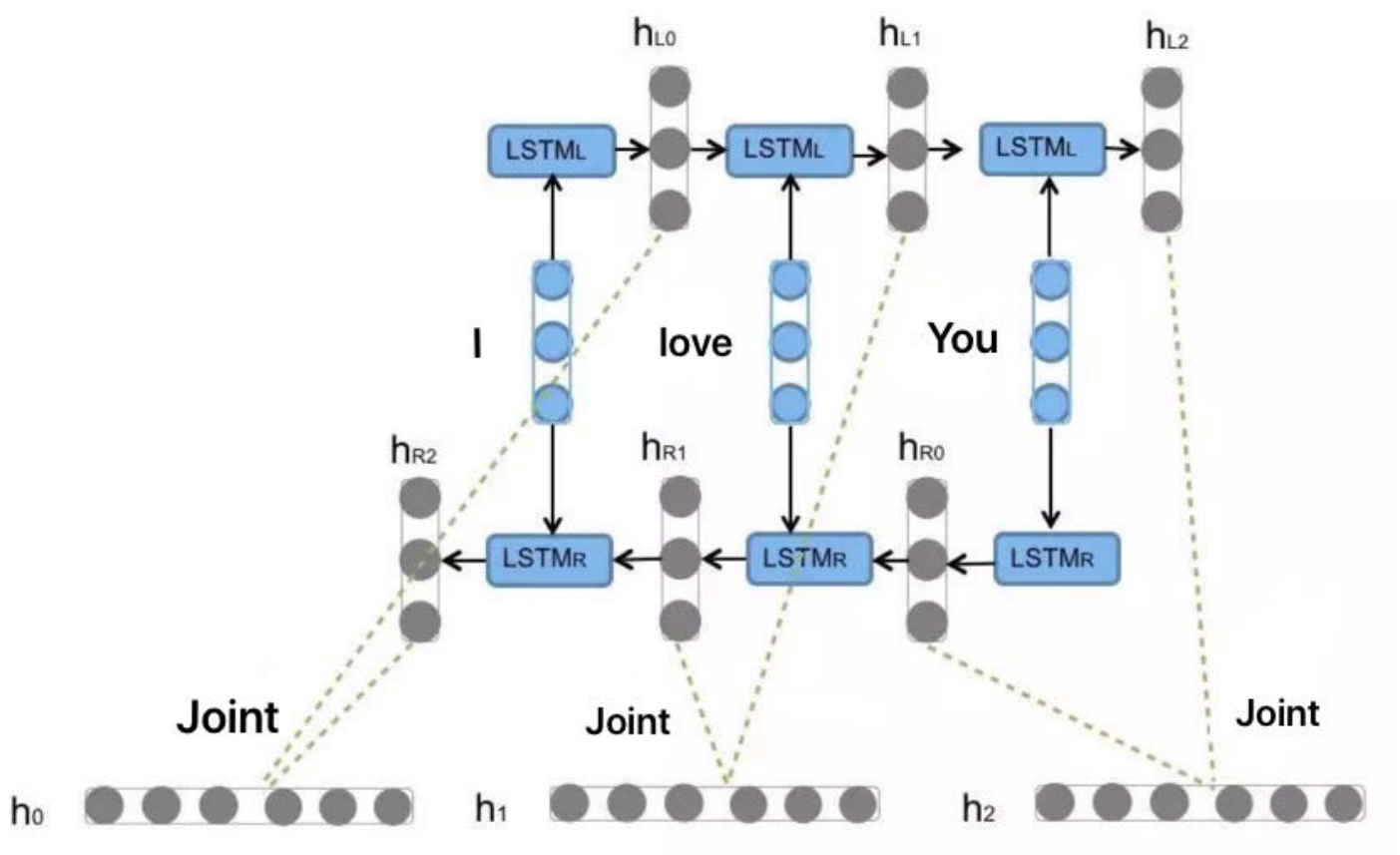

Figure 2.9: The principle of Bi-LSTM

**2.3.19 Albert**

**2.3.19.1 Bert**

The pre-training model Bert is the key experiment object of this project.BERT uses the Transformer Encoder model as the language model. The Transformer model completely discards RNN/CNN and other structures, and completely uses the Attention mechanism to calculate the relationship between input and output, as shown in the left half of the figure below, where the model Includes two sublayers:

Multi-Head Attention to do the Self-Attention of the model to the input

Feed Forward part to transform the input after the attention calculation

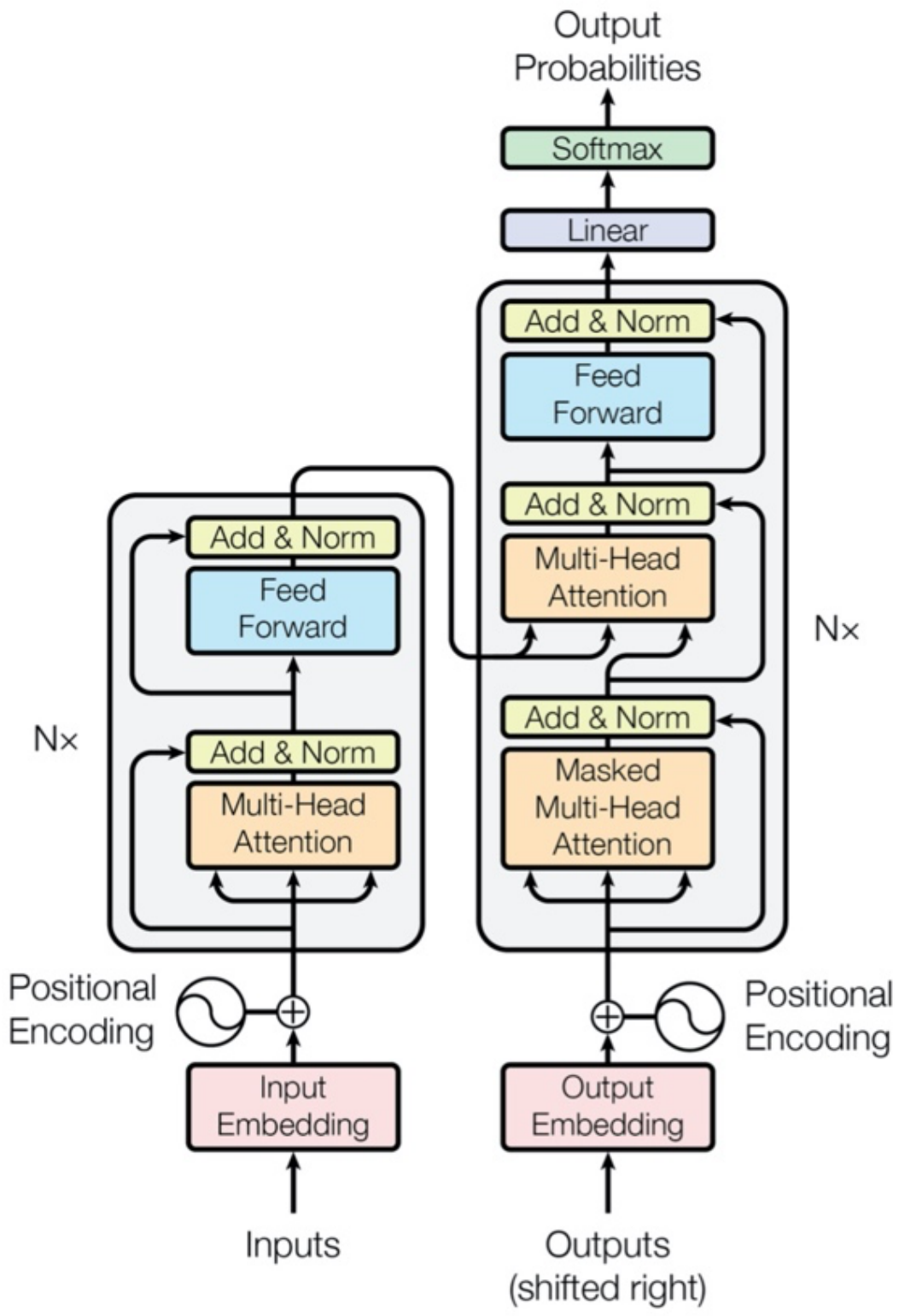

Figure 2.10: The structure of Transformer (Vaswani et al., 2017)

The BERT model is shown in the first on the left in the figure below. The difference between it and OpenAI GPT is that it uses Transformer Encoder, that is, the Attention calculation at each moment can get the input at all moments, while OpenAI GPT uses Transformer Decoder. The Attention calculation at a moment can only depend on the input at all moments before that moment, because OpenAI GPT uses a one-way language model.

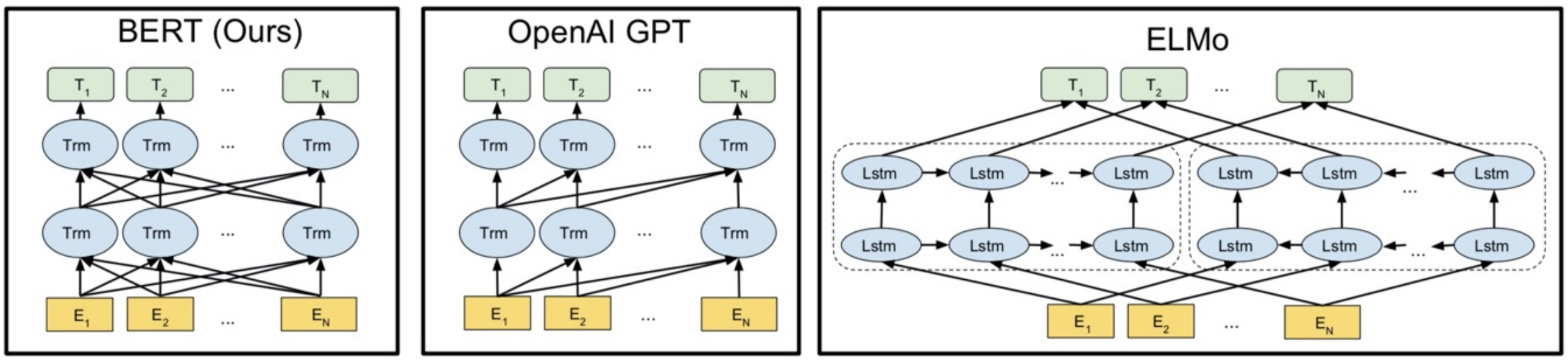

Figure 2.11: The difference between Bert, GPT and ELMo (Devlin et al., 2018)

**Embedding**

Bert's Embedding is made up of the sum of three types of Embedding.

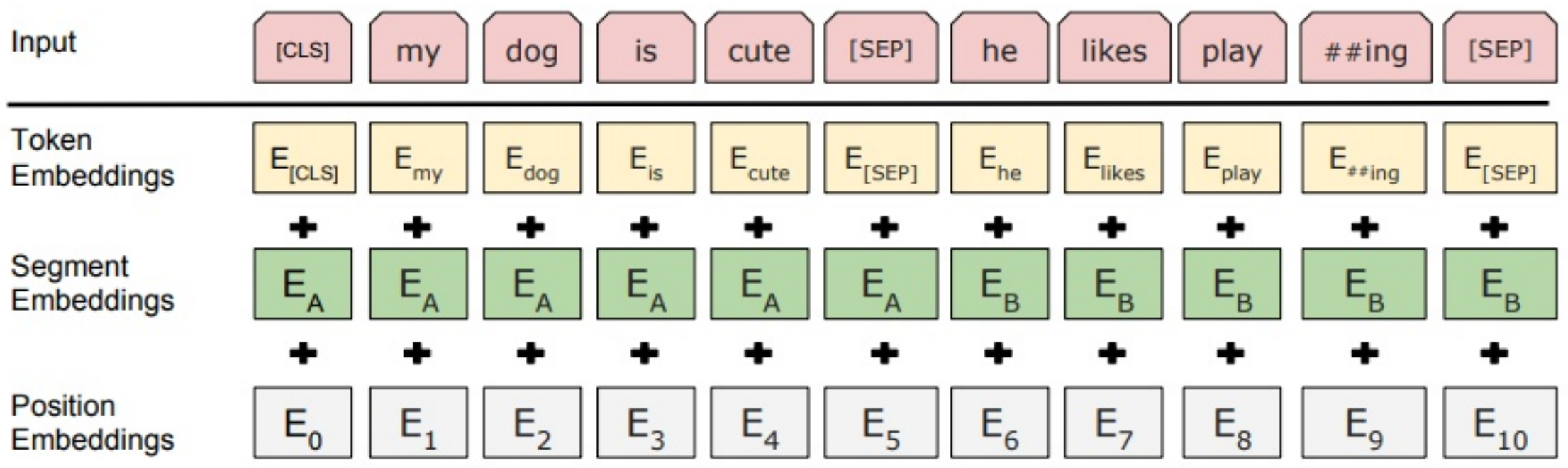

Figure 2.12: The embedding used in Bert (Devlin et al., 2018)

Token embedding represents the embedding of the current word. The first word is mask CLS, which is used for classification tasks.

Segment Embedding represents the index embedding of the sentence where the current word is located. Its purpose is to distinguish between two sentences, because pre-training not only builds a language model but also implements a classification task with two sentences as input.

Position Embedding represents the index embedding where the current word is located.

**Pre-training**

In Pre-training, BERT, in order to be able to benefit token-level tasks such as sequence label, and at the same time to facilitate sentence-level tasks such as question and answer, two pre-training tasks are used:

**Masked Language Model**

The goal of this pre-training step is to make a language model. From the model structure above, we can see the difference between Bert and other models, that is, bidirectional. If we use the pre-trained model to complete other tasks, we need contextual information, not just the information that follows.The model ELMo that takes this into account only trains left-to-right and right-to-left together. However, in fact, we actually want a deeply bidirectional model, but ordinary LM cannot do it. The author used a trick with mask.

In the training process, 15% of the tokens are randomly masked instead of predicting every word like CBOW. The final loss function only calculates the token that is masked. The method of Mask is not the same. Frequent use of [MASK] affects the performance of the model, so the author replaces 10% of the words in the random MASK with other words, keeps another 10% and replaces the rest with [MASK]. It should be noted that the pre-training model does not know in advance which words are covered, so you must pay attention to each word in the model. In addition, because the sequence length is too large (512), it will affect the speed of training, so 90% of the steps are trained with seq_len =128, and the remaining 10% of the steps are trained with 512-length input.

**Next Sentence Prediction**

In order to solve tasks such as QA and NLI, a second pre-training task was added to allow the model to understand the connection between two sentences. The training input is sentences A and B. There is a half chance of B being the next sentence of A. Input these two sentences and the model predicts whether B is the next sentence of A. During pre-training, 97-98% accuracy can be achieved.

**Fine-tuning**

Fine-tuning's modification of the model is very simple. For example, for emotion analysis task, the output of the pre-training model will as the input into the Softmax layer to get the classified output.In conclusion, different types of tasks require different modifications to the model, but the modifications are very simple, just add a layer of neural network. As shown below.

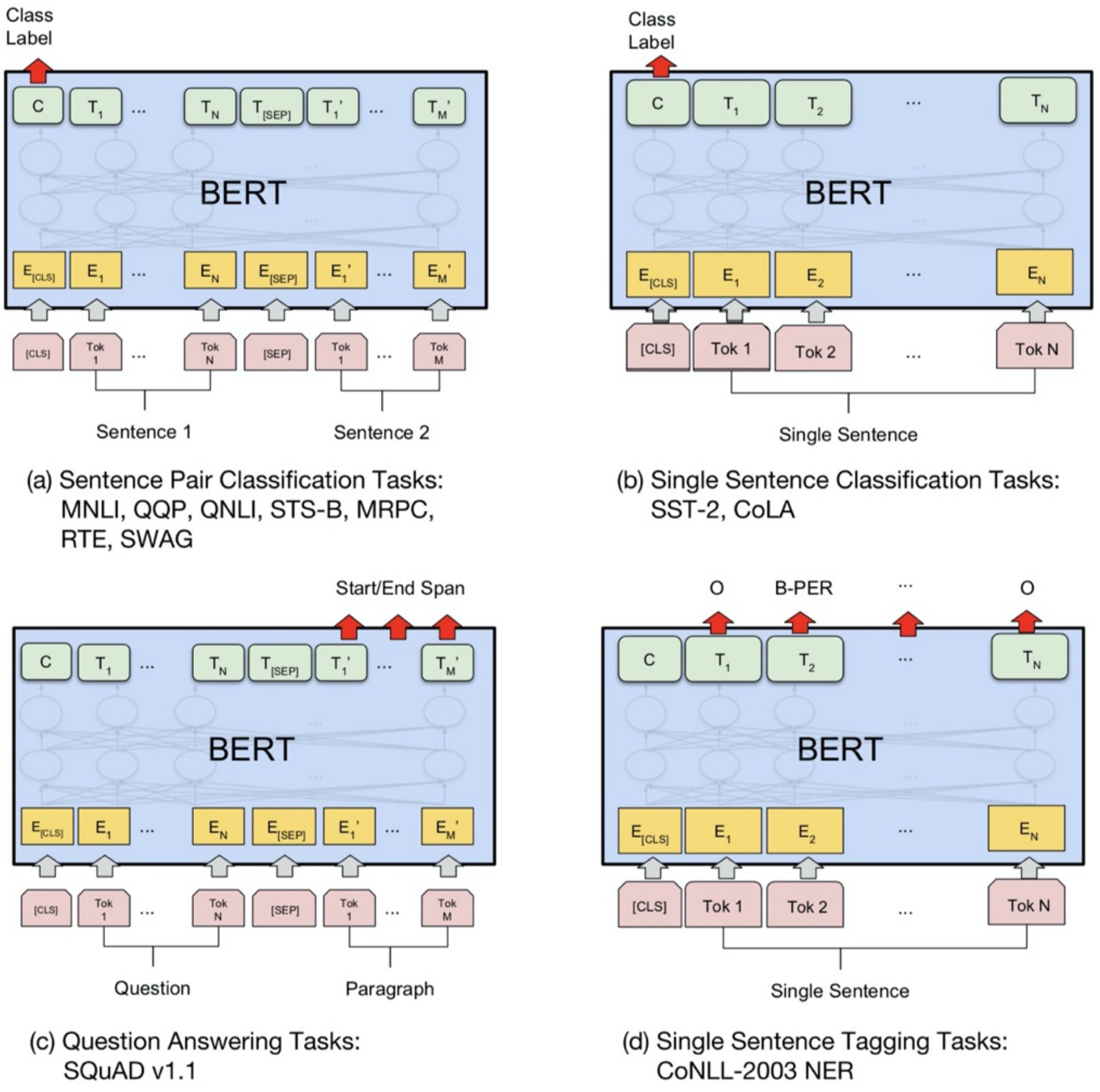

Figure 2.13: Fune-tune (Devlin et al., 2018)

#### 2.3.19.2 Albert Optimize

ALBERT mainly made three improvements on BERT, reducing the overall number of participants, speeding up the training speed and increasing the model effect.

**Factorized embedding parameterization**

BERT, XLNet and RoBERTa have the same embedding size(E) and hidden Size (H) of the transformer layer of the word table. This choice has two drawbacks:

1) From a modeling point of view, wordpiece vectors should be content-independent of the current content, while Transformer should learn representations that are content dependent. So separating E and H can make more efficient use of the parameters, because the H storing the context information is theoretically much larger than E.

2) From a practical point of view, vocab size in NLP task is inherently large. If E=H, the number of model elements will be easy to realize, and embedding training is sparse.

Therefore, the author USES a smaller E(64, 128, 256, 768) to train a context-independent embedding(VxE). Later, when computing, embedding into the hidden layer space (multiplied by a matrix of ExH) will be equivalent to doing a factor decomposition.

**Cross - layer parameter sharing**

Cross-layer parameter sharing means using only one Transformer for both 12 and 24 layers.

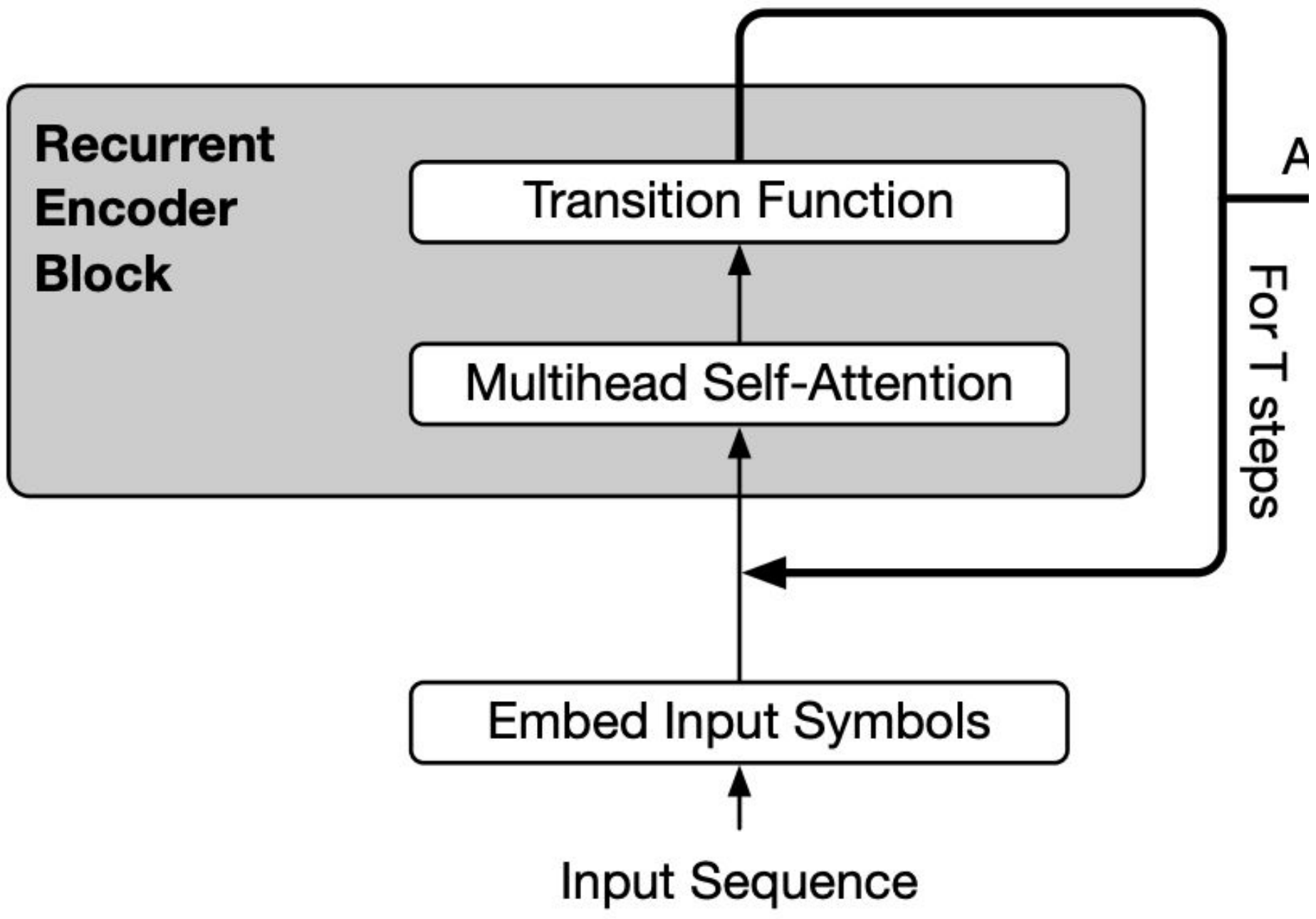

Figure 2.14: ALbert

The Albert's author compared the L2 distance and similarity of input and output of each layer, and found that BERT's results were quite oscillating, while ALBERT was very stable, so ALBERT has the role of stable network parameters.

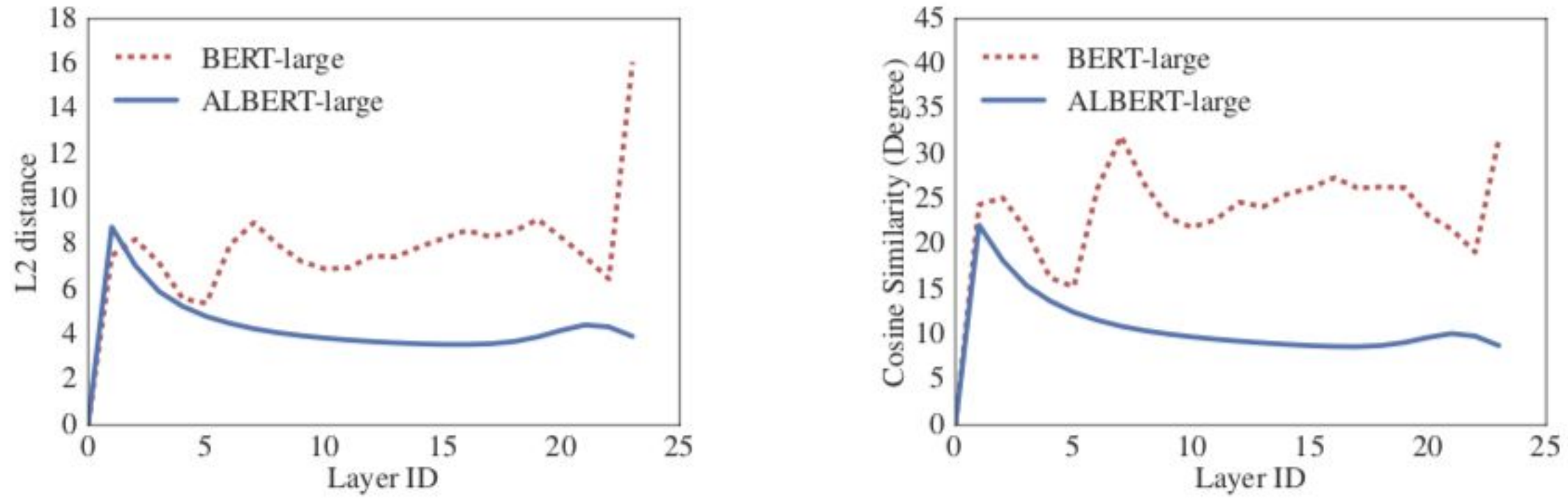

Figure 2.15: The compare between Bert and ALbert

**System - sentence coherence loss**

BERT, and a lot of research (XLNet, RoBERTa) were found the next sentence prediction useless, so ALbert also look at this problem, the author of that NSP is useless because the task not only contains the relationship between forecast, forecast also includes themes, theme and predict obviously simpler (such as a word from the news of finance and economics, in a word from the literary fiction), the model will tend to through topic related to predict. So replace it with a SOP to predict whether the sentence order has been swapped or not. The experiment shows that the new task has 1 point of improvement:

| | Intrinsic Tasks | | | Downstream Tasks | | | | | |
|---|---|---|---|---|---|---|---|---|---|
| SP tasks | MLM | NSP | SOP | SQuAD1.1 | SQuAD2.0 | MNLI | SST-2 | RACE | Avg |
| None | 54.9 | 52.4 | 53.3 | 88.6/81.5 | 78.1/75.3 | 81.5 | 89.9 | 61.7 | 79.0 |
| NSP | 54.5 | 90.5 | 52.0 | 88.4/81.5 | 77.2/74.6 | 81.6 | **91.1** | 62.3 | 79.2 |
| SOP | 54.0 | 78.9 | 86.5 | **89.3/82.3** | **80.0/77.1** | **82.0** | 90.3 | **64.0** | **80.1** |

Figure 2.16: The performance of different NLP tasks

## 2.4 Choice of Method

Although Weka is simple and easy to use. Compared to Python's rich libraries, it lacks ductility. The popular neural networks and deep learning frameworks in recent years are difficult to use. Moreover, WEKA runs in a fixed pattern and lacks flexibility. Therefore, this project chooses Python as the main tool to complete data preprocessing and model building evaluation.

For the model first layer word embedding, one-hot encoding is simple and easy to use, but it exists too thinly in vector space, wasting valuable resources. In this project, we use Word2vec for word embedding, which can save a lot of vector space and ensure that each feature has a unique vector. It is important to emphasize that N-gram has been apply in Albert, enabling it to understand the meaning of context and distinguish sentences. All of the eight algorithms mentioned in this chapter will be experimented to test their performance.

# Chapter 3
# Model Design

## 3.1 Overview

The objective of this project is to test the performance of different algorithms for offensive language recognition. A data set containing 14,200 posts is adopted, which is mainly divided into three subtasks: identifying whether the posts are offensive, determining whether the posts are offensive and what the objects are. For model establishment, it is divided into three parts: data preprocessing, modeling and evaluation. We adopted classical classification algorithms such as discrete Naive Bayes, KNN, Decision Tree, Random Forest, SVM, logistic regression as the base line algorithm, and Bi-LSTM and Pre-training model ALbert as the improved algorithm. This chapter mainly introduces the content of data set, the distribution and proportion of data, three sub-tasks and three components of the model.

## 3.2 Dataset

### 3.2.1 Introduction

Offensive language Identification Dataset (OLID) is a new large-scale dataset of English tweets with high-quality annotation of the target and type of offences. Previous work on offensive language identification has often targeted specific directions such as cyber violence, hate speech or cyberbullying. In contrast, the OLID dataset is designed as a whole, consisting of several different kinds of offensive content. In this dataset, using a hierarchical annotation schema split into three levels to distinguish between whether the language is offensive or not, categories of offensive language and its target (Zampieri er al., 2019).

**Level A: Offensive language Detection**

Offensive (OFF): Whether the post is targeted or not, it is offensive, such as racism, sexism, cyber violence, etc;

Not Offensive (NOT): Posts that do not contain offence or profanity.

**Level B: Categorization of Offensive Language**

Targeted Insult (TIN): Post that contains insulting aimed at individuals, groups or organizations;

Untargeted (UNT): Post that contains clearly insulting words but has no target.

**Level C: Offensive Language Target Identification**

Individual (IND): The post contains offensive remarks about a particular person. This could be a celebrity or a colleague or friend of the perpetrator. Such personal insults and violence are defined as cyberbullying.

Group (GRP): Posts contain insulting words aimed at a certain type of group. Targets can often be groups of ethnicity, political affiliation, sexual orientation, religious belief or other common characteristics. Such threats are often referred to as hate speech.

Other (OTH): The target of a post are usually not in the first two categories.

### 3.2.2 Analysis

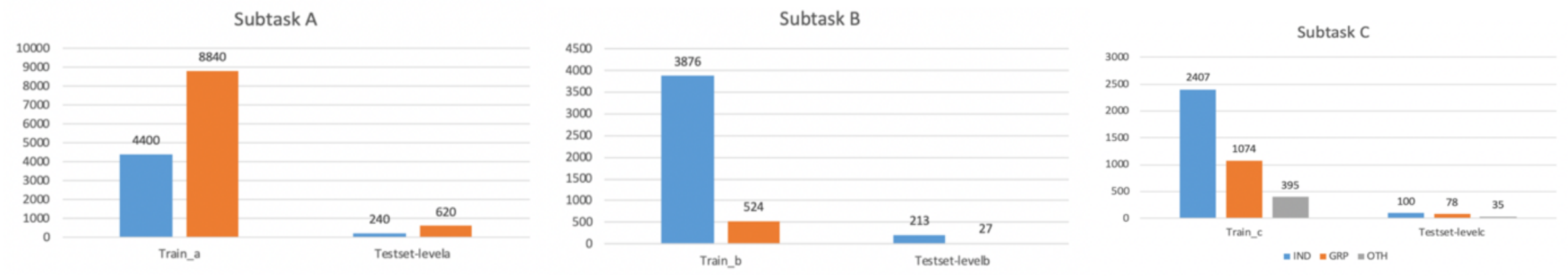

Figure 3.1: The proportion of data in different subtasks

As can be seen from the figure 3.1, the data set is heavily unbalanced. For subtask A, there are twice as many tags (NOT) as (OFF); In subtask B, there are six times more tweets targeted to an individual (TIN) than non targeted (UNT) ones; The number of classes in subtask C is not equal. (Batuhan Guler, 2019) Another problem is that there are Illegal words, sentences with incorrect grammar. Even native speakers still hard to avoid mistakes in writing, especially in situations such as impulsive writing. If the grammar of the sentence can be correct and the spelling of the word can be corrected, the data in the data set will be more accurate and the final classification/recognition result will be better.

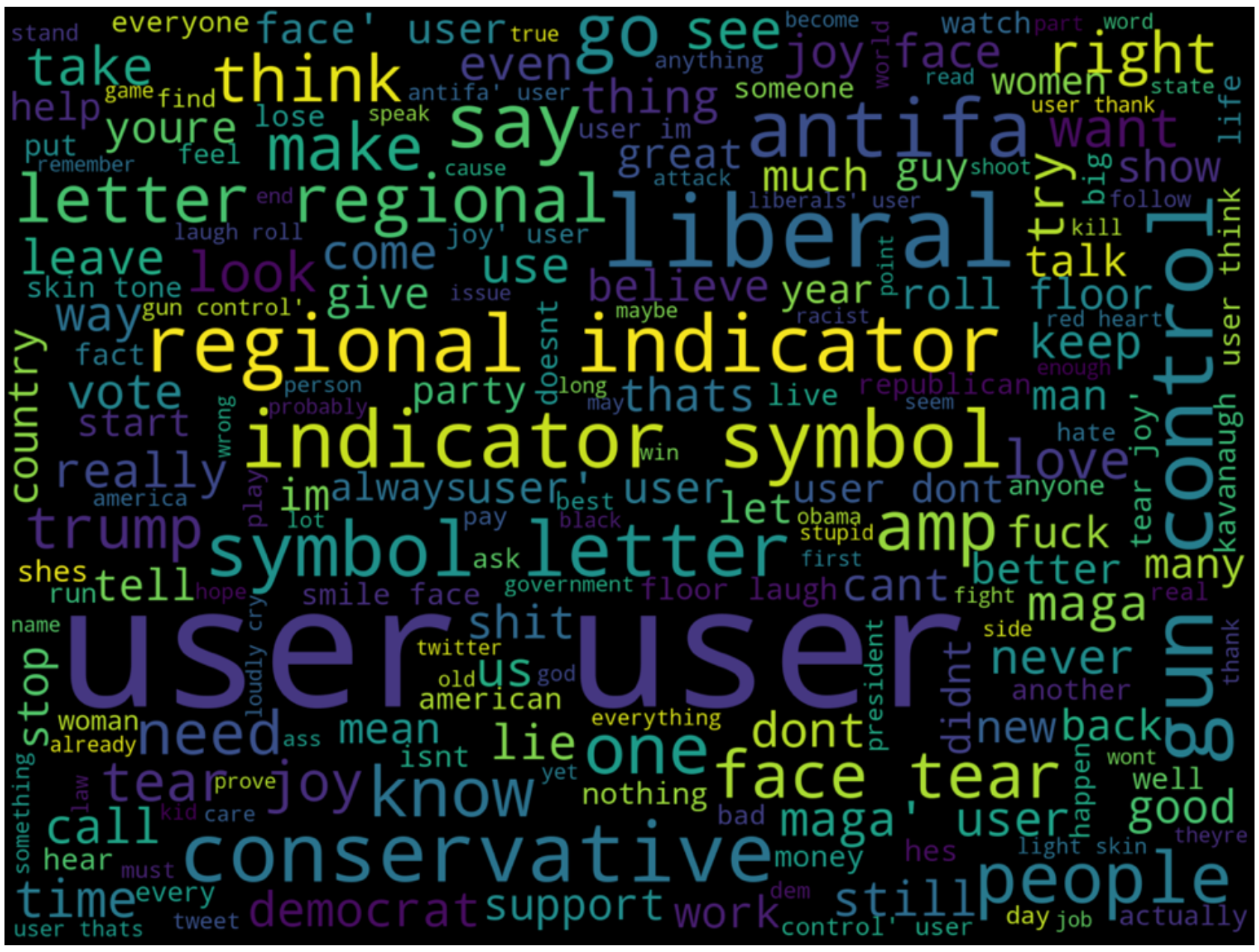

Figure 3.2: The frequency of word in dataset

This figure 3.2 shows the probability of each word appearing in the data set.

### 3.3 Tools

#### 3.3.1 The Python Programming Language

Python is a comprehensive scripting language that combines interpretability, interactivity and object orientation. Python is designed to be readable with similar to human language habits, and friendly to beginners.In this project, I chose python(3.7) based on the following factors.

1.Open source and portability: Python is one of FLOSS, which allows you to freely read and change its source code and use part of it in new free software. Base on its open source, Python has been ported to many platforms.These platforms include Linux, Windows, and Mac.

2.Scalability and mature package repositories: After 20 years of development, the Python standard library has become very large. It can easily help users to complete various tasks with high quality. In addition to the standard library, there are many other high-quality libraries, such as database processing, machine learning, deep learning, and Python image library.

3.Python is widely used in the field of artificial intelligence. Tensorflow, pytorch and other packages make the implementation of machine learning very simple.

#### 3.3.2 Deep Learning Framework

There are many deep learning frameworks available these days, most of them are open source. Figure 3.3 shows the most popular deep learning framework in the last six months of 2019. This project uses Keras to implement Albert model. The reason is as its documentation quote it: "Keras is an API designed for humans, not machines (Keras ,2015)." The Keras design is straightforward, which can reduce the workload of a large number of users, and is also easy to learn and use. Under this concept, even inexperienced users can quickly establish a deep learning model. At the same time, Keras does not sacrifice performance for simplicity. It integrates deep learning frameworks like CNTK and Tensorflow, making it highly flexible. As can be seen from the figure, TensorFlow and Keras are the two most popular deep learning frameworks in the past period.

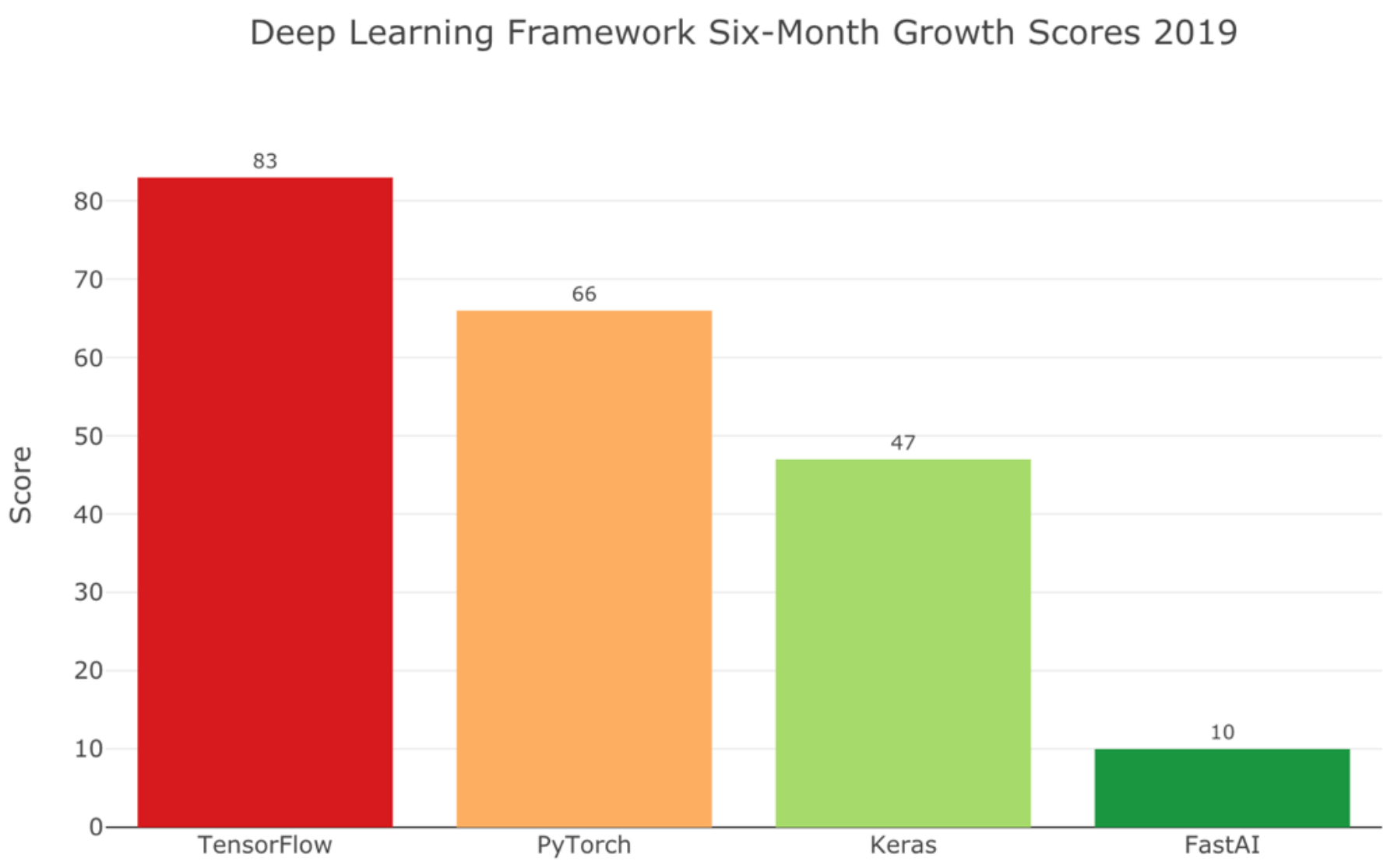

Figure 3.3: The population of Deep Learning Framework (Hale, 2019)

### 3.3.3 Nltk package

The natural language processing toolkit is a Python package commonly used in the field of NLP research. This is an open-source project that includes acquiring and processing a corpus, string processing, part-of-speech identification, etc. This expansion pack plays an important role in data preprocessing. In this project, the NLTK package is used to handle stop words, case and emove punctuation. It reduce us a lot of coding workload.

### 3.3.4 SKlearn package

Skleran is a very useful library in machine learning, covering algorithm and method from data preprocessing to training models, which saves users a lot of time in practice. This allows them to pay more attention to analyzing data distribution, adjusting models, and modifying parameters.In this project, SVM, Decision Tree, Random Forest, Logistic Tegression, Naive Bayes and KNN models are all established by sklearn. It is as simple as Keras, except that its library contains a large number of non-neural network models. Sklearn is fully functional, and we used it in our evaluation. Figure 3.4 shows the specific codes of evaluation indicators such as accuracy rate, F1-macro, survey and survey.

```
print("Accuracy", sk.metrics.accuracy_score(y_test, y_pred))
print("Precision", sk.metrics.precision_score(y_test, y_pred, average=None))
print( "Recall", sk.metrics.recall_score(y_test, y_pred, average=None))
print( "f1_score", sk.metrics.f1_score(y_test, y_pred, average=None))
print( "f1_score(macro)", sk.metrics.f1_score(y_test, y_pred, average='macro'))
print( "f1_score(micro)", sk.metrics.f1_score(y_test, y_pred, average='micro'))
print( "f1_score(weight)", sk.metrics.f1_score(y_test, y_pred, average='weighted'))
print( "MCC", sk.metrics.matthews_corrcoef(y_test, y_pred, sample_weight=None))
```

Figure 3.4: Using SKLearn to evaluation

### 3.3.5 Hyperparameter Tuning

When building the model, we set some hyperparameters for the model. Unlike normal parameters in neural networks, the value of the hyperparameter is set before the learning process begins. For neural networks, normal parameters are automatically updated by calculating weights and deviations. Although the model does not automatically adjust the hyperparameters set for it, such as adjusting weights and deviations, optimizing the hyperparameters is still a necessary step to achieve a good model. For the sklearn model, GridSearchCV is called. The function of GridSearchCV is to automatically adjust parameters. As long as the parameters are input, it will systematically go through a variety of parameter combinations and determine the best parameters through cross-validation.

## 3.4 Model Design

The aim of the project is to test the performance of various algorithms for offensive language identify. In order to achieve the goal of this project, the system design is divided into three subsystems that carry out separate and well-defined subtasks. As can be see from Figure 3.5, these subsystems are as follows: 1) text preprocessing 2) model establishment 3) comprehensive evaluation.

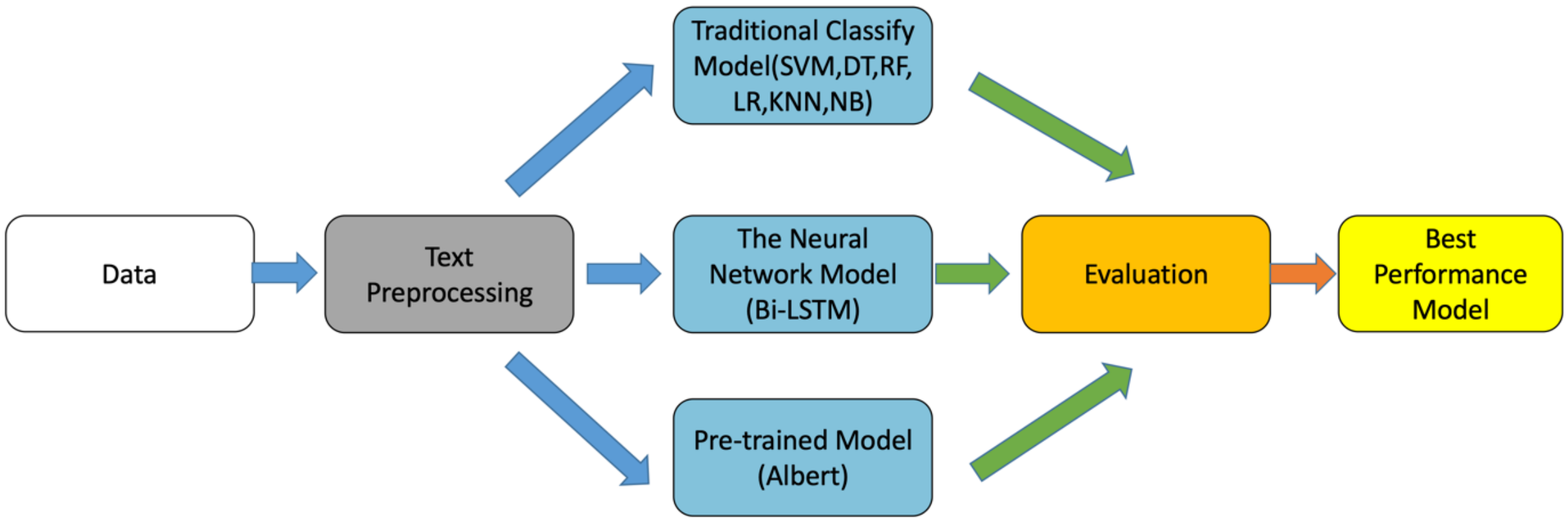

Figure 3.5: The Design of System

### 3.4.1 Text Preprocessing

Text preprocessing is an essential step in almost every natural language processing tasks. The purpose of the preprocessing is to make the text more predictable and analyzable by applying different text preprocessing techniques to transform the text into a unified form and remove the noise. Some complex natural language processing techniques, such as partial speech tags, are usually applied after a preprocessing task. In our system, we use the following several text preprocessing technologies:1) transforming emoji into text and 2) noise removal. The following sections discuss these techniques in detail.

**Transform emoji into text**

Most of the research has focused on identification methods. However, a good data set can effectively improve the performance of the algorithm. One potential direction for improvement is the pre-processing of emojis. In most researches, emoji were removed as noise. But emoji has become a new language that transcends the symbols themselves (Zhao et al., 2018). Translating emojis into their semantics (especially ironic emojis) can enhance the language model's understanding of context and improve its accuracy in identifying offensive posts. We translated the emoji into the corresponding text to make the context information more sufficient.

**Remove Noise**

Noise removal is one of the most basic and important text preprocessing techniques. Its main purpose is to eliminate characters, digits and symbols that may intrude with text analysis. Analyzing the text for the presence of noise may produce inconsistent results. Therefore, accurate noise removal is extremely important. We used the NLTK library and the open source code on GitHub to remove stop words, symbols, special characters, and convert all text to lowercase, expanding all abbreviations.

### 3.4.2 Algorithm Structure

The performance of each algorithm varies for different tasks because of the characteristics of each algorithm. For example, CNN usually gets better results than RNN when processing image recognition tasks, but worse results when processing text task. We used eight different algorithms for the experiment in order to find the best performance algorithm. Among them, SVM, decision tree, logistic regression, random forest, KNN, naive Bayes are called from the Sklearn library. LSTM and Albert are built by Keras library. Naive Bayes and SVM using special kind of model, which are described in detail as follows:

**Multinomial Naive Bayes**

Multinomial Naive Bayes algorithm used in this project, the classifier is mainly used for the type of text classification. It takes into account the number of occurrences of the word, which is term frequency.

**SVC**

The linear support vector classifier (SVC), which is a kind of SVM, uses linear boundaries to classify data. Compared with the nonlinear classifier, the complexity of the classifier is much lower and only suitable for small data sets. More complex data sets require a nonlinear classifier. The goal of a linear SVC is to match the supplied data, returning a "best-matched" hyperplane to divide or classify the data.

**Decision Tree and Random Forest**

As a good method of dividing the integrally unbalanced data set, we introduce them into the system for experiment. The specific calculation method of cross entropy will be determined according to the automatic parameter adjustment mentioned in Section 3.3.5

**KNN**

For data sets lacking prior probabilities, KNN will be the first choice. However, there is a imbalance in the data set used in this project, and the weight of each K will be affected. Therefore, it will be an important experimental to observe the performance of KNN on this dataset.

**Logistic Regression**

Although regression models are used more for regression tasks than classification tasks, testing the performance of regression models in classification tasks will provide a reference for future improvements.

**Bi-LSTM**

The performance of Bi-LSTM has always been excellent at processing text tasks. It can remember the sequence information in both directions, which is helpful to understand the meaning of the context, which is very helpful to identify the offensive in the sentence.

**Albert**

Through extensive text training in advance, ALbert achieved the best results in 11 NLP tasks. We used training data to fine-tune it and apply it to this project.

### 3.5 Evaluation Measures

By unified evaluation, the performance of different algorithms will be objectively compared and measured. This will help us find good algorithms to save time for future work. Acc as the most basic measurement method has its limitations, so we added F1-Macro and MCC to evaluate the performance of unbalanced data sets. In addition, confounding matrix is also considered because it can well evaluate multiple indicators.

### 3.5.1 Acc

Accuracy is the most common and easy-to-understand evaluation index, which is the number of divided samples divided by the number of all samples. Generally, the higher the accuracy, the better the performance of the classifier.

$$Accuracy = \frac{True\ positive + True\ negative}{True\ positive + False\ positive + True\ negative + False\ negative} \quad (3.1)$$

Accuracy is indeed an excellent and intuitive evaluation index, but sometimes high accuracy does not mean good performance of the model. For example, in the case of imbalanced data, the accuracy of this evaluation indicator has great flaws. For example, in the case of imbalanced data, the accuracy of this evaluation indicator has great flaws. For expamle, In earthquake prediction, the probability of an earthquake occurring is small, typically a few parts per thousand or less. If the accuracy rate is used as the evaluation criterion, even if all the predictions will not happen, the accuracy rate still will be greater than 99%, but the earthquake early warning system will not play any role in the event of an earthquake. Therefore, it is far from scientific and comprehensive to evaluate an algorithm model only based on accuracy.

### 3.5.2 Confusion matrix

The confusion matrix is an important tool to evaluate classification models. It not only intuitively expresses the performance of the model, but also extends a large number of evaluation indicators, such as precision, recall and F1 scores. In a confounding matrix, each row represents the prediction, each column represents the reality, and the values on the diagonal are the correctly classified values., so that ROC curves can be drawn, and the performance of classification models can be more comprehensively evaluated and compared.

The role of confusion matrix in this project:

1) It is used to observe the performance of the model in identifying each category, and calculate the accuracy and recall of the model corresponding to each category;

2) Through the confusion matrix, we can observe which categories are not easy to distinguish. For example, how many categories of offensive posts are classified as non-offensive posts, so that targeted design features can be used to make the categories more distinguishable.

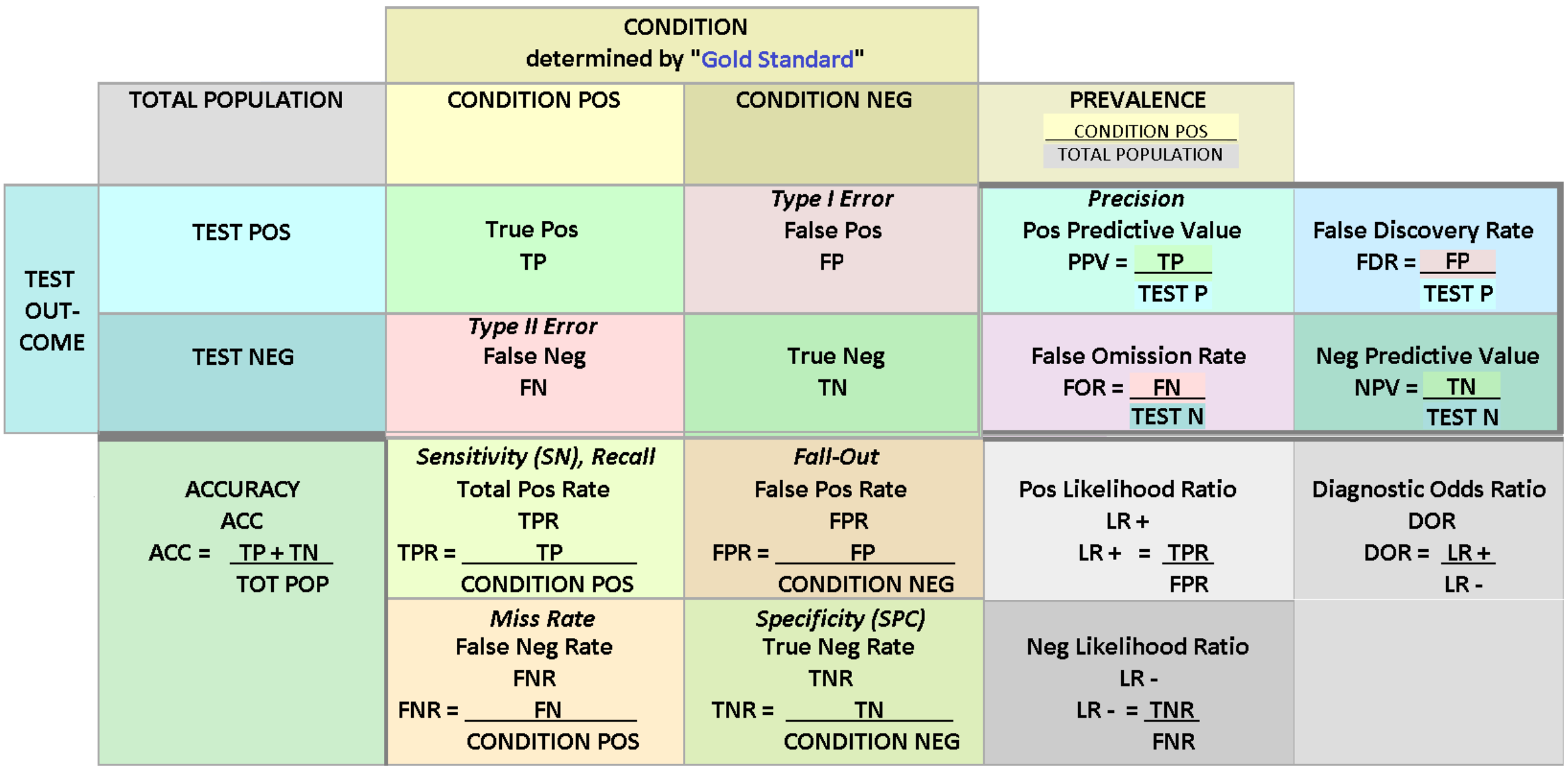

Figure 3.6: Confusion Matrix （Lavender888000）

### 3.5.3 Recall

Recall rate, which refers to the original sample, means the probability of being predicted as a positive sample in an actual positive sample.

$$Recall = \frac{True\ positive}{True\ positive + False\ negative} \tag{3.2}$$

### 3.5.4 Precision

Precision is the probability of a positive sample of all the samples that are predicted to be positive. In other words, how certain are we that we are going to be correct in predicting the outcome of a positive sample. Precision and accuracy look similar, but they are completely different concepts. Precision represents the prediction accuracy in the positive sample results, while accuracy represents the overall prediction accuracy, including both positive and negative samples.

$$Precision = \frac{True\ positive}{True\ positive + False\ positive} \tag{3.3}$$

### 3.5.5 F1

In some cases, we need to maximize precision or recall at the expense of another metric. For example, in a preliminary screening test that follows patients, we might expect a recall rate close to one -- we want to find all the patients who actually have the disease. If the cost of follow-up examination is not very high, we can accept lower accuracy. However, if we want to find the best combination of accuracy and recall, F1 Score is the best choice.

$$F1 = \frac{2 * Precision * Recall}{Precision + Recall} \tag{3.4}$$

F1 Score is a harmonic average of precision and recall. The reason for using harmonic average rather than simple arithmetic average is that harmonic average can punish extreme cases. A classifier with a precision of 1.0 and a recall rate of 0 would have an arithmetic average of 0.5 for both indices, but an F1 score would be 0. F1 Score, which gives the same weight to recall and precision, is a special case of the universal F β metric, in which β can be used to give more or less weight to recall and precision.

### 3.5.6 MCC

MCC is a comprehensive indicator mainly used to measure multi-classification problems. Due to the comprehensive consideration of TP TN, FP and FN, the effect is better than F1 in the case of unbalanced data samples. The value range of MCC is [-1, 1]. A value of 1 means that the prediction is completely consistent with the actual result; 0 means that the prediction result is similar to the random prediction result; -1 means that the prediction result is completely inconsistent with the real result. Therefore, MCC essentially describes the correlation coefficient between the predicted results and the actual results.

$$MCC = \frac{TP * TN - TP * FN}{\sqrt{(TP + FP) * (TP + FN) * (TN + FP) * (TN + FN)}} \tag{3.5}$$

# Chapter 4
# Detailed Model Design

## 4.1 Overview

The purpose of the model is to realize the recognition of offensive language. According to the design and tools of the Chapter 3, this chapter mainly discusses the concrete methods of implementing data preprocessing, the structure and parameter setting of the model, finally complete the implementation of the model.

## 4.2 Data preprocess

### 4.2.1 Emoji substitution

We use one emoji code on Github which could map the emoji Unicode to substituted phrase. We translate emoji into English phrases to maintain their semantic meaning. Especially when the size of dataset is limited, this can expand it. The Firgure 4.1 shows how it transforms an emoji into a corresponding meaning based on Unicode.

```
>> import emoji
>> print(emoji.emojize('Python is :thumbs_up:'))
Python is 👍
>> print(emoji.emojize('Python is :thumbsup:', use_aliases=True))
Python is 👍
>> print(emoji.demojize('Python is 👍'))
Python is :thumbs_up:
>>> print(emoji.emojize("Python is fun :red_heart:"))
Python is fun ❤
>>> print(emoji.emojize("Python is fun :red_heart:",variant="emoji_type"))
Python is fun ❤ #red heart, not black heart
```

Figure 4.1: The example of emoji substitution

### 4.2.2 Remove the excessive URL

The users mentioned in twitter are anonymous and appear as @user to protect their privacy. However, if we know the specific content of @user, we can easily determine whether the object is an individual or an organization. For example, “A @FoxNews commentator just ripped me with lies, with nobody defending.@edhenry”.It's easy to know from looking up tweets, where @FoxNews is an organization and @edhenry is an individual. Unfortunately, the OLID dataset anonymizes all @ users, making it impossible to know the type of object being insulted that way. So we Delete the excessive @users, so that each tweet has a maximum of 3

@users, for a Consecutive '@user's are limited to three times to reduce the redundancy and ensure that the integrity and semantics of the sentence does not change.

```
def take_data_to_shower(tweet):
    noises = ['URL', '@USER', '\'ve', 'n\'t', '\'s', '\'m']

    for noise in noises:
        tweet = tweet.replace(noise, '')

    return re.sub(r'[^a-zA-Z]', ' ', tweet)
```

Figure 4.2: Code for remove Noise

### 4.2.3 Convert all the context to lower

Capitalized words appear less frequently in a given data set, although they are more offensive. Instead of making the data more difficult to process, we decided to convert all letters to lowercase to reduce noise.

```
def tokenize(tweet):
    lower_tweet = tweet.lower()
    return word_tokenize(lower_tweet)
```

Figure 4.3: Code for convert lower letter

### 4.2.4 Non-ASCII filtering

Twitter contains two types of non-ASCII characters: emojis and non-Latin characters. For the former, we have translated it into English phrases. For non-Latin characters, as OLID is a data set intended to contain only English tweets, hence these characters can be discarded as noise.

### 4.2.5 HashTag segmentation

Converts hashtag to text, such as' # LunaticLeft is segmented as' the Lunatic Left which is obviously offensive in this case. This can greatly improve the text information.

### 4.2.6 Convert the full-width to half-width

We do this to ensure that the format is uniform.

### 4.2.7 Stop word removed

Since stop words contibute little meaning to the recognition of tweets, we deleted them.

### 4.2.8 Punctuation removed

All punctuation such as question marks, colons, etc. are removed to reduce noise.

### 4.2.9 Apostrophe handling

Tweets usually do not require strict compliance with writing requirements, and there are many oral forms of expression. This will cause noise and hides negation which is important for offensive detection. So we convert colloquial abbreviations into written forms. For example, "it's" into "it is; "can’t" into "can not".

```
def replace_abbreviations(text):
    texts = []
    for item in text:
        item = item.lower().replace("it's", "it is").replace("i'm", "i am").replace("he's", "he is").replace("she's", "she is")\
            .replace("we're", "we are").replace("they're", "they are").replace("you're", "you are").replace("that's", "that is")\
            .replace("this's", "this is").replace("can't", "can not").replace("don't", "do not").replace("doesn't", "does not")\
            .replace("we've", "we have").replace("i've", " i have").replace("isn't", "is not").replace("won't", "will not")\
            .replace("hasn't", "has not").replace("wasn't", "was not").replace("weren't", "were not").replace("let's", "let us")\
            .replace("didn't", "did not").replace("hadn't", "had not").replace("waht's", "what is").replace("couldn't", "could not")\
            .replace("you'll", "you will").replace("you've", "you have")
        item = item.replace("'s", "")
        texts.append(item)
    return texts
```

Figure 4.4: Code for apostrophe handling

### 4.2.10 HTML removed

Some HTML entities in OLID dataset such as &gt have no specific meaning in tweets, so we remove it.

### 4.2.11 Lemmatisation

Tokenization is the process of demarcating and possibly classifying sections of a string of input characters. In this process, they are lots of methods that could be used to make a different type of tokens such as a lemmatized token, stemming token and regular token. Lemmatization and stemming are two important ways of word normalisation. Stemming is the extraction of the stem or root form of a word (which does not necessarily express the full meaning). Compare with stemming, lemmatization relies on a wider range of contexts such as paragraphs or entire documents to correctly predict the exact part of speech and meaning of a word.The regular tokens we used for our coursework convert to lowercase only for all words, removing the punctuation. However, lemmatization will convert all the words to the base form. For example, the word 'went' will be converted to 'go'.

The main advantage of lemmatization is that it has a better tolerance for noise and it takes into consideration the meaning of the word context to predict its meaning, which means that its performance will be better and more efficient. In addition, converting words with the same root back to the prototype can effectively increase the data set. For example, in this project, we delete the word that appears less than three times. The absence of these words leads to the loss of title meaning. If we use the lemmatization, then these occurrences fewer words are likely to merge with other word, to ensure accuracy.

### 4.3 Word embedding

Use code to convert words into word vectors. Select any three dimensions of the trained word vectors and display them in the coordinate system. It will be found that semantically similar words will be very close to each other in the spatial coordinates, while semantically irrelevant words will be far apart from each other. This property can be used for more general analysis of words and sentences (Tang, 2014).

### 4.4 Modeling

#### 4.4.1 KNN

I set a variety of parameters for the remaining five models and got the best parameters with the ability to automatically tune GridSearchCV. The parameter n_neighbors in the KNN algorithm selects 3,5,7,9 to represent the number of neighbors. 'uniform' in weights means that the hyperparameter of distance weight is not considered, while 'distance' means that the hyperparameter of distance weight is considered.

#### 4.4.2 SVC

In SVC, C is the penalty parameter. The larger C is equivalent to penalizing the relaxation variable, hoping that the relaxation variable is close to 0, that is, the punishment for misclassification increases and tends to be in the case that the training set is fully divided into two pairs. In this way, the test accuracy of the training set is high, but the generalization ability is weak. A small C value reduces the penalty for misclassification, allows fault tolerance, treats them as noise points, and has strong generalization ability. I set C to 0.001, 0.01, 0.1, 1,10. The best penalty parameter was determined by GridSearchCV.

#### 4.4.3 Logistic Regression

Logistic regression has more parameters. First is multi class classification parameters choice, namely optional parameters for ovr and multinomial, which I set to automatically use the default parameters. Second is solver: choose parameter optimization algorithm, only five optional parameters, namely the Newton - CG, LBFGS, liblinear, sag, saga. The default is liblinear. Solver parameters determine our optimization method for logistic regression loss function. In this project, Newton-CG, the second derivative matrix of loss function, namely Heisen matrix, is used to iteratively optimize the loss function. Third, penalty: The specification used to specify the penalty term, with optional arguments of L1 and L2. The L1G specification assumes that the parameters of the model meet The Laplace distribution, while L2 assumes that the model parameters meet the Gaussian distribution. The so-called normal form is to add constraints on parameters to make the model less likely to overfit. Here we choose L2 because the Newton-CG solver only supports the L2 specification. Finally, C: inverse of the regularization coefficient. Like SVM, smaller values indicate stronger regularization.

```
elif(type=="LR"):
    classifier = LogisticRegression(multi_class='auto', solver='newton-cg',)
    classifier = GridSearchCV(classifier, {"C":np.logspace(-3,3,7), "penalty":["l2"]}, cv=3, n_jobs=4)
    classifier.fit(train_vectors, train_labels)
    classifier = classifier.best_estimator_
```

Figure 4.5: Code for Logistic Regression

#### 4.4.4 Decision Tree and Random Forest

For decision tree and random forest, due to the large data sample size and many features, I set the maximum depth of both of them as 800 and the minimum sample number of each node as 5. At the same time, both methods of criterion, Gini and entropy, are taken into account. The number of decision trees in the random forest is set between 50 and 200.

#### 4.4.5 Multinomial Naive Bayesian

For the Multinomial Naive Bayesian model, After many tests, I chose the $\lambda$ value as 0.7.

#### 4.4.6 Bi-LSTM

1.For the LSTM model, the key point is that the embedding layer must be defined first. The model must ensure that the input text data is converted into word vectors that the model can compute. In our model, the Max feature of this layer is set to 2000, and the embed size is 256.

2. After defining the embedding layer, it is a major part of the model-LSTM layer. The shape of the batch input must be defined, but the activation function is optional. In addition, the Bidirectional function is added to this layer to enable the LSTM to learn the sequence from the back to the forward.

3. After the LSTM layer, the output must conform to the label to give the result. The Dropout layer is added to randomly discard a portion of the data to avoid overfitting. After this is the full connection layer as the classifier. Defining the last dense layer with Softmax enables the prediction of the most likely classes by setting the maximum probability to 1 and the others to 0 to accommodate the first thermal coding of the original tag. The number of neurons in the last layer set to 2 means that it will be divided into two categories.

4. When all the layers are defined, we must define the model itself and compile it. In the compile API, we can set other super parameters, such as optimizer, evaluation index, etc.

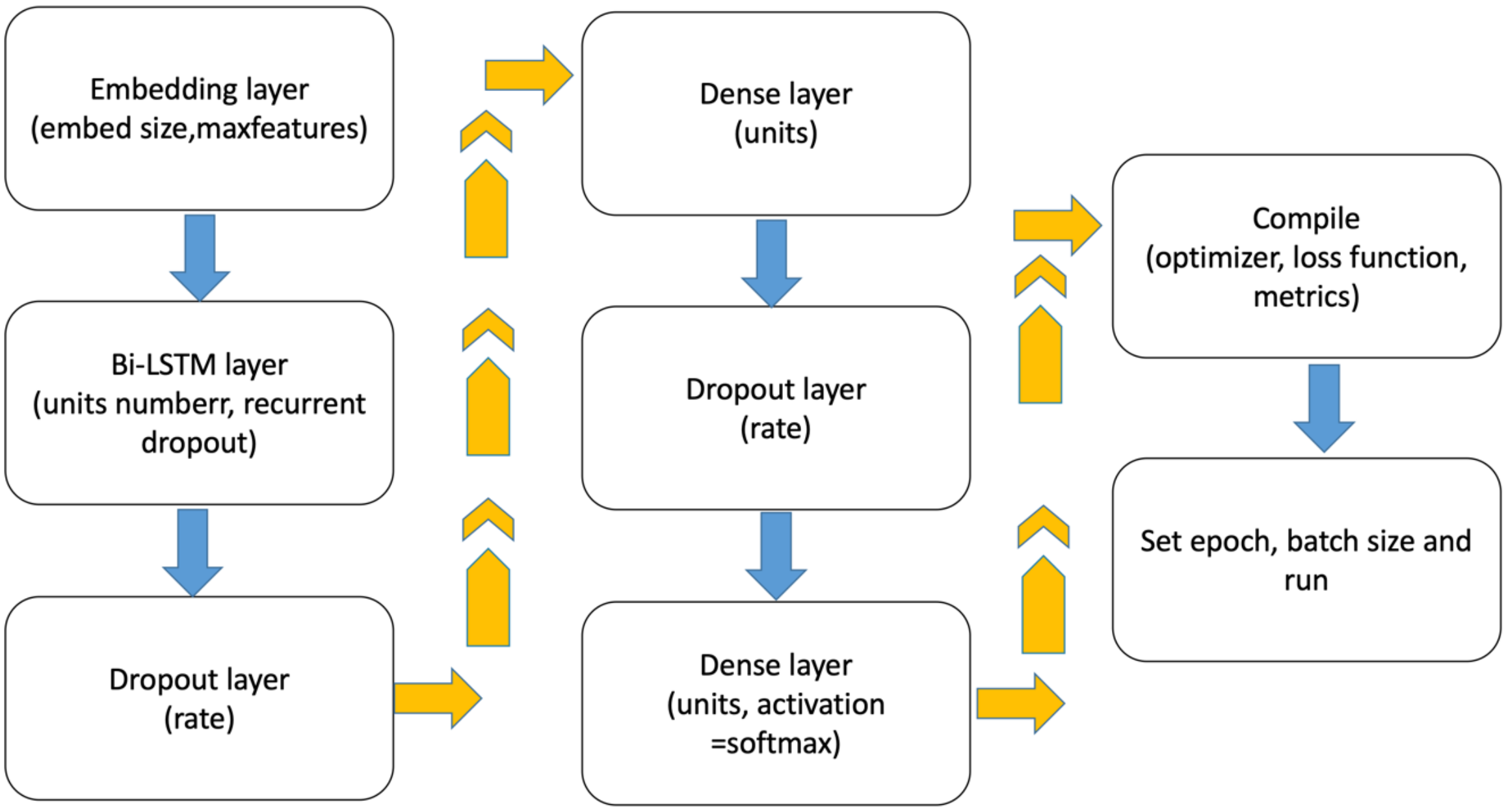

Figure 4.6: Structure for Bi-LSTM

An epoch means that an entire data set is loaded into a neural network at once, so the number of epochs represents the number of times a neural network has been trained to master data and patterns. Just like a student needs to learn the same word many times before he/she can say it correctly. Neural networks usually cannot master all the features at one time, and more time is needed to study the same data to master their features and patterns. So theoretically, with epochs increases, the model's performance will better. In this project, we carried out multiple epochs of sub-tasks A and B respectively, and the results are shown in the figure below:

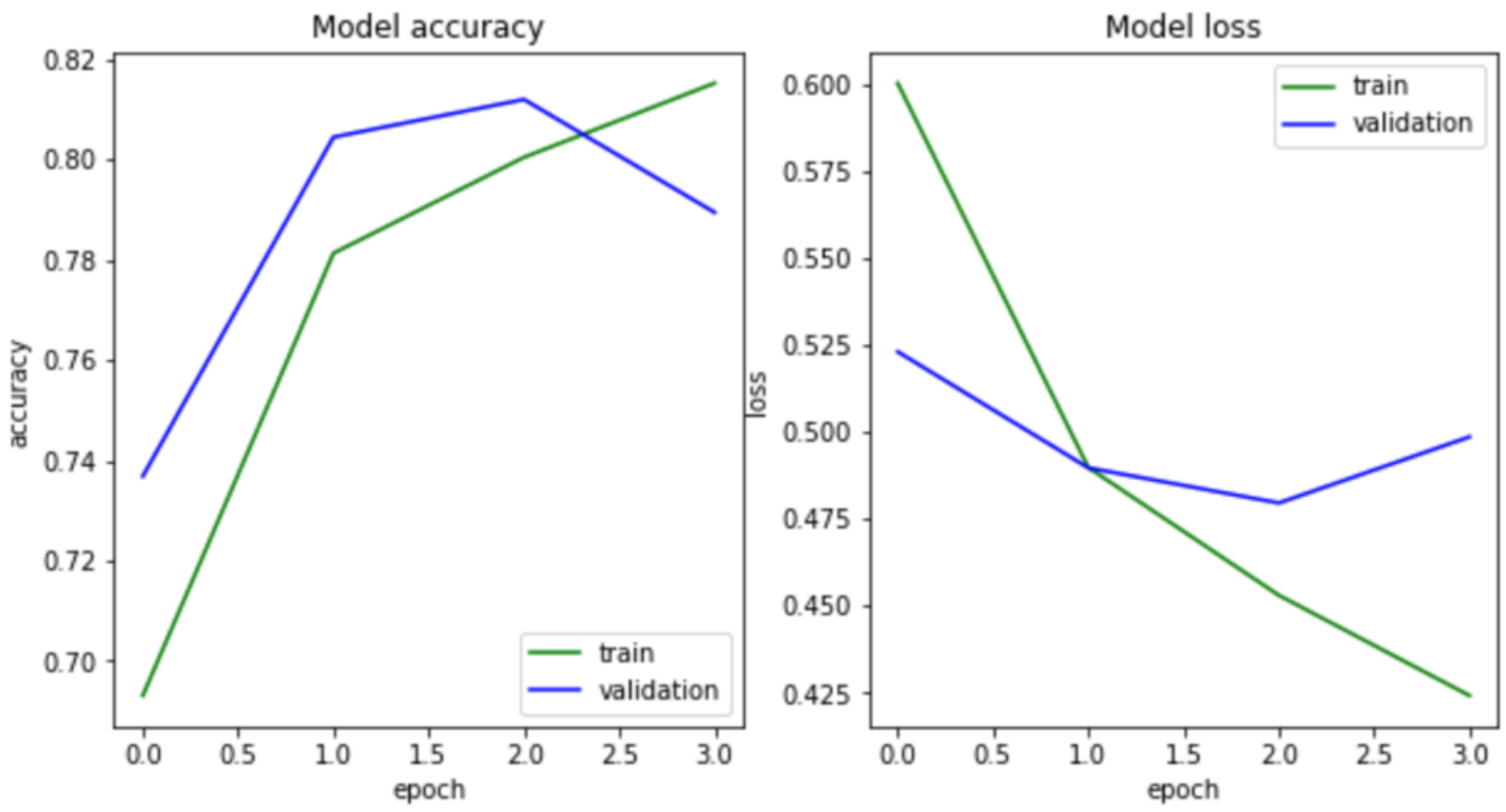

Figure 4.7: Accuracy and Loss of Bi-LSTM in subtask-A

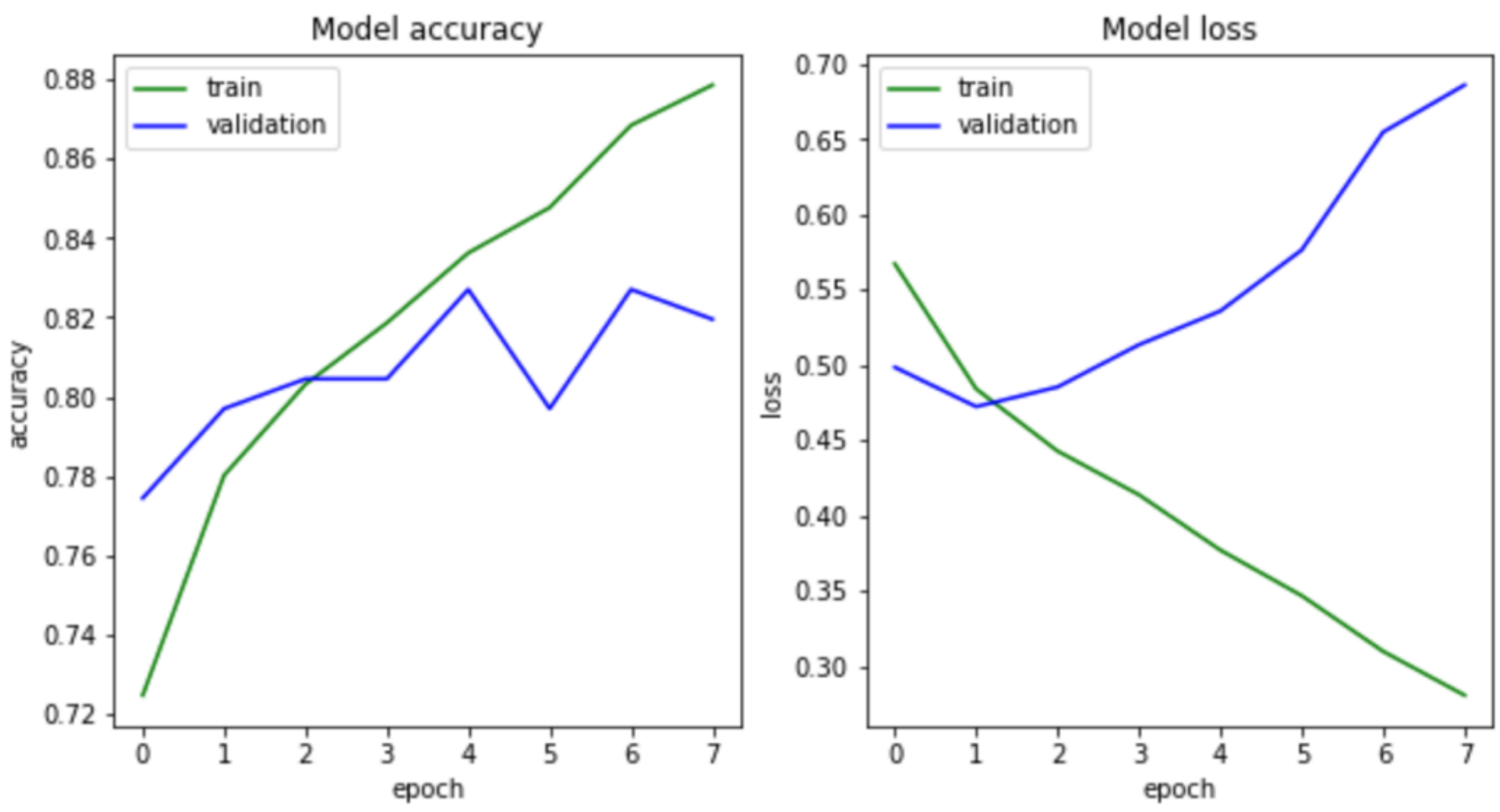

Figure 4.8: Accuracy and Loss of Bi-LSTM in subtask-B

Overfitting refers to the generalization of the model which is very poor. The trained model performs well for the training set but poorly for the test set. As can be seen from the figure, after a epoch, training Loss continued to decline, but verification loss began to rise, which meant that the model had overfitted. Therefore, we set the number of epochs to 1.

Table 4.1: Batch Size Experiment Setup

| Batch Size | Accuracy |
|---|---|
| 2 | 0.6756 |
| 8 | 0.6847 |
| 16 | 0.7592 |
| 32 | 0.8320 |
| 64 | 0.8210 |

If the data set is too large to be passed it into the model at once, we need to divide the data set into batches. In programming, this is present by setting the size of the batch, which means how many examples we want to load in a batch. In fact, we call the process of loading a batch is an iteration. However, iteration is not just a loading process. After each calculation of the model, the error function is used to calculate the gradient descent, carry out back propagation, and update its weight and deviation. Therefore, the number of iterations also means the number of self updates of the network during a training period. We set different batch sizes in Table 4.1, with the number of epochs being 1. The results are shown in Table 4.1. Interestingly, accuracy does not always increase with batch size, and it has been proven that performance degrades when the batch size exceeds a certain point. The reason behind this is that because of the number of iterations, when the batch size increases the number of iterations decreases, the weight and bias are not updated as frequently as they should be, affecting the outcome of learning, just as a student learning too much in a day may result in remembering less.

| Layer (type) | Output Shape | Param # |
|---|---|---|
| embedding (Embedding) | (None, None, 256) | 512000 |
| bidirectional (Bidirectional | (None, 64) | 73984 |
| dropout (Dropout) | (None, 64) | 0 |
| dense (Dense) | (None, 64) | 4160 |
| dropout_1 (Dropout) | (None, 64) | 0 |
| dense_1 (Dense) | (None, 2) | 130 |

Figure 4.9: Structure of Bi-LSTM

Stackingi more layers to the neural network means that each sample can be extracted based on features, thus improving the performance of the model. However， as with the number of epochs, there is a trade-off between the time it take to run and the number of layers. Due to time constraints, this project did not have the opportunity to test the performance of more BI-LSTM layers. The specific performance of this model will be shown in chapter 5.

### 4.4.7 Albert

The author has adjusted the model to a better state, so instead of improving the structure of the model, we applied fine-tune. Different from BI-LSTM, training Albert was time-consuming, we ran 35 epochs at a time and stored the best parameters for the experiment.

```
def on_epoch_end(self, epoch, logs=None):
    val_acc = evaluate(valid_generator)
    if val_acc > self.best_val_acc:
        self.best_val_acc = val_acc
        model_name = '../models/best_model_%s.weights' % self.f_key
        model.save_weights(model_name)
        logging.info('model save to:%s' % model_name)
    test_acc = evaluate(test_generator)
    logging.info(
        u'val_acc: %.5f, best_val_acc: %.5f, test_acc: %.5f\n' %
        (val_acc, self.best_val_acc, test_acc)
    )
```

Figure 4.10: Get the best parameters

For best performance, we chose Albert-XXLarge, whose parameters are shown in the figure above. The reason for reducing the number of layers from 24 to 12 is that the performance does not improve significantly as the number of layers increases, but the time consumed increases significantly (Zhang et al., 2019).

| Model | | Parameters | Layers | Hidden | Embedding | Parameter-sharing |
|---|---|---|---|---|---|---|
| BERT | base | 108M | 12 | 768 | 768 | False |
| | large | 334M | 24 | 1024 | 1024 | False |
| ALBERT | base | 12M | 12 | 768 | 128 | True |
| | large | 18M | 24 | 1024 | 128 | True |
| | xlarge | 60M | 24 | 2048 | 128 | True |
| | xxlarge | 235M | 12 | 4096 | 128 | True |

Figure 4.11: The configurations of the main BERT and ALBERT models(Zhen et al.,2019)

1. This function gives data token.

2. Running with 35 epoches, we used the data in OLID to fine-tune Bert to make its performance more suitable for the classification tasks of this project. The final optimal parameters will be used for subsequent experimental evaluation. We also need compile in each fine-tuning. In compile, we set the optimizer and loss function to update the parameters.

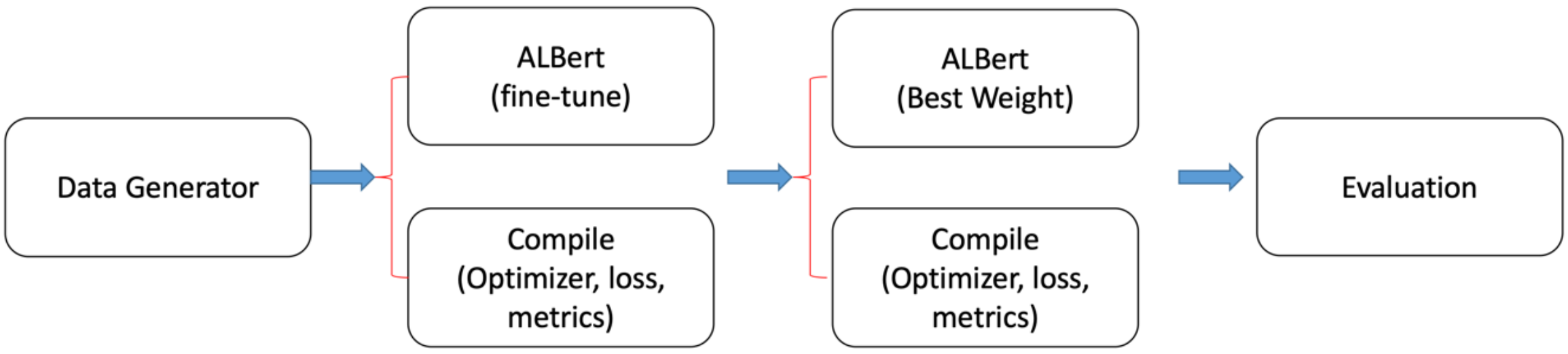

Figure 4.12: The process for train and test Albert

# Chapter 5
# Result and Evaluation

## 5.1 Overview

This chapter summarizes the experimental results and analyzes the performance of the algorithm in each task. In addition, we explain the reasons for some phenomena: this is mainly due to the nature of algorithms and the fact that data sets are unbalanced.

## 5.2 Result

### 5.2.1 Data Preprocessing

In order to achieve the goal of removing noise from data, 11 kinds of data preprocessing were performed. Figure 5.1 shows the state of the training set before preprocessing. There are "#", "@user", emoji, capitalization, non-English symbols and so on. Figures 5.2, 5.3 and 5.4 show the effect after preprocessing. Posts become text-only and most of the noise is removed or modified.

| | |
|---|---|
| 96874 | #RAP is a form of ART! Used to express yourself freely. It does not gv the green light or excuse the behavior of acting like an animal! She is not in the streets of the BX where violence is a w |
| 65507 | @USER Do you get the feeling he is kissing @USER behind so he can humiliate him later? |
| 78910 | 5 Tips to Enhance Audience Connection on Facebook URL @USER #socialmedia #smm URL |
| 46363 | #BiggBossTamil janani won the task. She is going to first final list👏👏👏👏 |
| 68123 | #Conservatives - the party of low taxation 😂 #Tories #Tory URL |
| 22452 | 𝓒-𝓒𝓮𝓭𝓻𝓪𝓷.. The Nord cannot make a single move, but he is fully aware of what is happening. A chill comes down his spine as a ghostly, rather. . . calm female voice called for him from |
| 15565 | #ConsTOO THE PLACE FOR FED UP CONSERVATIVES !!! |
| 64376 | #GreatAwakening #QAnon #PatriotsUnited #WWG1WGA #AreYouAwake WHEN YOU ARE AWAKE YOU SEE CLEARLY ❤️🧡💛💚 Check out this video URL URL |

Figure 5.1: Data of trainset before Preprocessing

Moreover, we also made three training sets according to different tasks. Figure 5.2 shows the pre-processed training data of subtask A, in which the label '1' means non-offensive and the label '0' means offensive. Figure 5.3 shows the pre-processed training data of subtask B, where the label '1' represents the existence of certain objects and the label '0' represents the absence of specific offensive objects. Figure 5.4 shows the pre-processed training data of sub-task C, in which the label '2' represents the offending object is an individual, the label '1' represents the offending object is a group, and the label '0' represents the offending object does not belong to the previous two categories.

| | | |
|---|---|---|
| 154 | boltey hai na ki jo beef kaate hai uski akal gutno me reti hai n i can expect it brain he is the pm of my country n yes we celeb his bday every day n it make sense the ppl who offer prayers fr osama | 1 |
| 155 | got something to hide eric | 1 |
| 156 | anderson lies and you report it as fact you do get except for blm amp antifa you have no real viewers | 1 |
| 157 | niggaz buying pussy bitches trying to buy love | 0 |
| 158 | you are so correct jin is a whole crackhead | 1 |
| 159 | thats right walk away from democrats  vote red to save america  votered2018  mcga maga | 1 |
| 160 | amazing all the projection she is doing | 1 |
| 161 | reasonable gun control  use 2 hands | 1 |
| 162 | antifa members have to be their 1 clients | 1 |
| 163 | at this point its just a money grab theyre not even trying to hide it now | 1 |
| 164 | he went broke and got propped up by white liberals in hollywood thats white privilege all day long stop defending coke head tom arnold | 1 |
| 165 | made it last night for you  rolling on the floor laughing  rolling on the floor laughing  rolling on the floor laughing | 1 |
| 166 | and no one is attacking trump supporters at large  ever heard of antifa  i thought you were leaving long ago anyway  why are you still wasting my time | 1 |
| 167 | you are being super cute with your hormone storys and i hope all 3 of you are well | 1 |
| 168 | consider the source its only bono | 1 |
| 169 | fuck those terrorists in the ass | 0 |
| 170 | i think she is lying about that too  i have seen no threats and i have seen a thousand comments and articles on this story | 1 |
| 171 | because president trump and god gave us a miracle | 1 |
| 172 | she was drunk this rape is her fault  he was drunk this rape isnt his fault  said no one except the crying liberals making more stuff up you guys are laughable | 1 |
| 173 | carrey is becoming weird like peewee herman | 0 |
| 174 | i love the way leona alleslev has seen through the liberals she has undoubtedly seen the destruction of our choice of service to our country trudeaus father started its destruction in the 60s and it ha | 1 |
| 175 | tried to comment  on his journalistic smear he made justice clarence thomas but for some reason he has blocked 68yr old christian gmother who tweets occasionally when she sees the injustice of | 1 |
| 176 | crush him beto  make him look like the idiot he is | 0 |
| 177 | go home youre drunk  maga trump2020  oncoming fist  regional indicator symbol letter u  regional indicator symbol letter s  oncoming fist | 0 |
| 178 | a super super generalised statement what you are saying goes against every economic meteric one of the biggest reasons irisheuropean gdp has increased is because of freedom of movement if y | 1 |
| 179 | i hope they beat his ass in private | 0 |
| 180 | the interview is a very important part of the hiring process and selling yourself may not come naturally make sure youre fully prepared  tn | 1 |

Figure 5.2: Data of subtask A after Preprocessing

| | | |
|---|---|---|
| 258 | what a freaking ass | 0 |
| 259 | way to go david thumbs up  i love it when someone can send the liberals into a tessyfiy in the morning keep up the fight  regional indicator symbol letter u  regional indicator symbol letter s  regional indicator symbol | 1 |
| 260 | its funny how you idiot liberals seem to forget when other presidents do the same shit haha | 1 |
| 261 | hiding like the caged animal he is | 1 |
| 262 | they embraced antifa career killer | 1 |
| 263 | these cowards aint ready maga  regional indicator symbol letter u  regional indicator symbol letter s | 1 |
| 264 | gets period you are the cause of my | 1 |
| 265 | by which point the fuckingclass was nearly over lol | 1 |
| 266 | start by not tweeting pussy ass shit | 1 |
| 267 | black heart  dyrus you are a fucking worthless braindead scumfuck bastard pile of trash mental dickface that should be gunned down in the street like the degenerate you are black heart | 1 |
| 268 | landon sucks i asked him for my birthday if i could play with him but he never got back to me so landon you suck | 1 |
| 269 | i like how her boobs have the same mass as her belly xd | 1 |
| 270 | what the fuck | 1 |
| 271 | good luck with that i dont follow along with fascists like the resist movement and their apple and google overlords i know what happened in ww2 and its the rhetoric they use that put us there i dont need a god to see | 1 |
| 272 | john podesta is a spirit cooking pedo heres a photo of him and valerie jarrett in a rather compromising situation  maga q a non drain the deep state drain the swamp | 1 |
| 273 | thanks joe my sanity was forsaken a looong time ago hahaha | 1 |
| 274 | you are name calling a decorated disabled veteran  shame on you know your subject before making slanderous remarks | 1 |
| 275 | same tbh could they just stop spreading rumors and false news we are hurt enough we dont need all this shit | 1 |
| 276 | and guys like you are the fat yokels | 1 |
| 277 | face with tears of joy  face with tears of joy  face with tears of joy  face with tears of joy  face with tears of joy  liberals really are nuts | 1 |
| 278 | i have heard he is making waves | 1 |
| 279 | this jag off ref tony corrente needs to get out of the nfl he is a bias piece of garbage he made a horrible bad call in packer game which changed the game he is a liar and wont admit when hes wrong nfl please fire to | 1 |
| 280 | democrats gun control and mass shootings data determined to be fraudulent  democrats are liars once again | 1 |
| 281 | she is getting death threatscan you go and be a shield for her so you dont have to wait | 1 |
| 282 | choosing heartless politician over your ancestors country  speaks volumes as to who you are | 0 |
| 283 | no just imagining if we found a picture of bret holding a woman like this liberals will go insane | 1 |
| 284 | he is delirious | 1 |
| 285 | this woman has nothing but play tennis and yet she is constantly being disrespectful its insane the whole all owes her an apology atp | 1 |

Figure 5.3: Data of subtask B after Preprocessing

| | | |
|---|---|---|
| 150 | is it gun control or anti 2a  i havent seen a single ad attacking 2a  thats just  rhetoric to create fear | 1 |
| 151 | so alyssa you saying hes guilty  what ever happened to innocent before guilty  and if he is innocent would you at least tweet an apology  shut up and answer the tweet | 2 |
| 152 | trumps lies have grown far more frequentand more dangerous  mother jones  truth decay maga were doomed | 2 |
| 153 | conservatives have become the cadaver party | 1 |
| 154 | fuck a drug test | 2 |
| 155 | bizarre story and more to it once she is charged formally glad she is off the streets and certainly concerned if there are more clueless cops like her out there protecting us | 2 |
| 156 | my friend is involved in the gun control movementhad the same discussion with her she really believes semi automatic guns aka  assault rifles spray bullets with one trigger pull total ignorance | 2 |
| 157 | gop on personal responsibility  he was drunk boys will be boys  she was drunk slut  republicans conservatives kavanaugh | 2 |
| 158 | turn them loose and see what happens to her city people like her is what gives the democrats a bad name she is risking the entire population of atlanta wonder how the citizens like it well see at th | 1 |
| 159 | bono is an idiot | 2 |
| 160 | according to the liberals yes  guilty unless she proves her innocence  wait are you a man if so then no you can not be trusted ever how dare you make a claim against a woman women are above | 1 |
| 161 | you are that mean ass cousin we all hate lol | 2 |
| 162 | he is what the industry stands for all about spreadsheets and no care for the human side | 2 |
| 163 | i am so weak she really beat her ass | 2 |
| 164 | hubby rushed me shakes fist will flick a pic thru this evening when i feed em again by the way he is still the boss | 2 |
| 165 | down with liberals | 0 |
| 166 | you are a bad ass keep fighting for us  two hearts | 2 |
| 167 | you are such a lier and all for money nobody wants  francoism here in spain except ppl like you and your fascists separatists friends to smoke screen all the money embezzlement here in cataloni | 2 |
| 168 | just please dont let cruz get away with the he is a tax and spend liberal racist dog whistle crap fire back with facts and call him out on bs | 2 |
| 169 | nope with mm no waymaybe is time to ask sky what has gone wrong because foster from cnn have the book first than kp tweeted about the book  liberals from trump for americans to mm for britis | 1 |
| 170 | all 20 of them  hardly a problem until the antifa of hundreds showed up  for what   to riot and destroy | 1 |
| 171 | soooooounds like an awesome plan   next hes gonna be pulling money from medicare amp pensions to pay for all the golf cart rentals for his  detail that watches him struggle to stay on the fairwa | 2 |
| 172 | oh wow you look like a 1980s porn star loveit | 2 |
| 173 | republicans grow some balls and start playing dirty like these liberals assholes do | 1 |

Figure 5.4: Data of subtask C after Preprocessing

### 5.2.2 Model Result

This section presents the results of the evaluation of various algorithms in three sub-task. The following chart shows the performance of each algorithm in subtask A. In subtask A, Albert achieved the best performance with an accuracy 79.59% and F1-Macro 0.9227. Due to the unbalanced data set, MCC was added to this project for evaluation. The results show that Albert also performs best among all the tested algorithms. When we come to recall, all algorithms performs better in one class and worse in another class. Albert have the better performance than other algorithms in precision, which OFF class(99.60%) and NOT class(83.05%). Bi-LSTM and Decision Tree also achieved the great performance in the NOT class (82.58%) and OFF class (78.89%) respectively.

Table 5.1: Result of Subtask A

| System | Accuracy | Class | Precision | Recall | F1-macro | MCC |
|---|---|---|---|---|---|---|
| Decision Tree | 72.71% | OFF | 78.89% | 81.19% | 0.6850 | 0.3707 |
| | | NOT | 58.80% | 55.26% | | |
| Naive Bayes | 72.35% | OFF | 71.44% | 97.41% | 0.5879 | 0.3235 |
| | | NOT | 81.31% | 22.42% | | |
| KNN | 68.12% | OFF | 68.07% | 98.18% | 0.4755 | 0.1551 |
| | | NOT | 69.46% | 8.22% | | |
| Random Forest | 77.55% | OFF | 77.26% | 93.97% | 0.7093 | 0.4658 |
| | | NOT | 78.78% | 44.74% | | |
| Logistic Regression | 75.16% | OFF | 74.08% | 95.58% | 0.6676 | 0.4188 |
| | | NOT | 81.10% | 36.20% | | |
| SVM | 76.16% | OFF | 77.22% | 91.06% | 0.7007 | 0.4311 |
| | | NOT | 72.29% | 46.47% | | |
| Bi-LSTM | 80.58% | OFF | 72.12% | 49.58% | 0.7205 | 0.4519 |
| | | NOT | 82.58% | 92.58% | | |
| ALbert | 79.59% | OFF | 99.60% | 85.85% | 0.9227 | 0.8405 |
| | | NOT | 83.05% | 99.50% | | |

In subtask B, Albert still performs best (accuracy 91.42%, F1-Macro 0.9816. This was followed by Bi-LSTM, whose accuracy is close to that of Albert. On the other hand, Naive Bayes, Logistic Regression and SVM all show the phenomenon that only one category can be classified, among which the recall of one category is 100% and that of the other category is 0. This indicates that these classifiers are not suitable for sorting tasks in unbalanced data sets.

Table 5.2: Result of Subtask B

| System | Accuracy | Class | Precision | Recall | F1-macro | MCC |
|---|---|---|---|---|---|---|
| **Decision Tree** | **84.72%** | **UNT** | **90.23%** | **92.82%** | **0.5757** | **0.1545** |
| | | **TIN** | **27.08%** | **20.96%** | | |
| **Naive Bayes** | **87.90%** | **UNT** | **87.90%** | **100%** | **0.4678** | **0.0000** |
| | | **TIN** | **0%** | **0%** | | |
| **KNN** | **68.12%** | **UNT** | **68.07%** | **98.18%** | **0.4755** | **0.1551** |
| | | **TIN** | **69.46%** | **8.22%** | | |
| **Random Forest** | **88.09%** | **UNT** | **88.94%** | **98.87%** | **0.4969** | **0.0572** |
| | | **TIN** | **26.66%** | **3.22%** | | |
| **Logistic Regression** | **87.09%** | **UNT** | **87.09%** | **100%** | **0.4655** | **0.0000** |
| | | **TIN** | **0%** | **0%** | | |
| **SVM** | **87.45%** | **UNT** | **87.45** | **100%** | **0.4665** | **0.0000** |
| | | **TIN** | **0%** | **0%** | | |
| **Bi-LSTM** | **90.00%** | **UNT** | **57.89%** | **40.74%** | **0.5884** | **0.1731** |
| | | **TIN** | **92.76%** | **96.24%** | | |
| **ALbert** | **91.42%** | **UNT** | **99.68%** | **96.68%** | **0.9816** | **0.7757** |
| | | **TIN** | **65.62%** | **95.45%** | | |

In subtask C, both SVM and Bi-LSTM perform best. Only in GRP class, Bi-LSTM has higher precision(72.85%) and recall (65.38%) than SVM (25% and 2.32% respectively). Unlike tasks A and B, ALbert performed poorly on the multiclassification task, even being the worst performer of all the algorithms, with an accuracy rate of 70.48%.

Table 5.3: Result of Subtask C

| System | Accuracy | Class | Precision | Recall | F1-macro | MCC |
|---|---|---|---|---|---|---|
| Decision Tree | 60.78% | IND | 62.06% | 37.50% | 0.4267 | 0.3096 |
| | | OTH | 69.10% | 92.25% | | |
| | | GRP | 50.00% | 1.14% | | |
| Naive Bayes | 63.67% | IND | 60% | 4.51% | 0.2866 | 0.1004 |
| | | OTH | 63.75% | 99.18% | | |
| | | GRP | 0% | 0% | | |
| KNN | 63.98% | IND | 54.16% | 14.33% | 0.3407 | 0.1649 |
| | | OTH | 64.80% | 96.34% | | |
| | | GRP | 50.00% | 1.05% | | |
| Random Forest | 68.62% | IND | 65.26% | 38.79% | 0.4273 | 0.3261 |
| | | GRP | 0% | 0% | | |
| Logistic Regression | 67.80% | IND | 62.06% | 37.5% | 0.4267 | 0.3096 |
| | | OTH | 69.10% | 92.25% | | |
| | | GRP | 50% | 1.14% | | |
| SVM | 70.48% | IND | 56.70% | 43.30% | 0.4505 | 0.3369 |
| | | OTH | 74.44% | 90.77% | | |
| | | GRP | 25% | 2.32% | | |
| Bi-LSTM | 70.89% | IND | 67.69% | 88% | 0.4694 | 0.3398 |
| | | OTH | 92.30% | 34.28% | | |
| | | GRP | 72.85% | 65.38% | | |
| ALbert | 51.42% | IND | 50% | 1.92% | 0.0370 | 0.0751 |
| | | OTH | 50% | 0.81% | | |
| | | GRP | 51.42% | 99.45% | | |

### 5.3 Evaluation

We can conclude that in subtask A, there is not a big difference in the performance of all kinds of algorithms. One potiential reason is that there are enough data to train the model. The training set for subtask A has 13,000 posts and there is not much significant difference of proportion between the two classes. Nevertheless, we can still find that all algorithms have a certain category of lookup significantly higher than the other, indicating that even if the data set imbalance is not obvious, it will affect the algorithm's recognition of the target.

The performance of various models in subtask B is different, among which the traditional SVM algorithm and logistic regression algorithm both recognize a category of 0%. There are two reasons for this phenomenon :1) In this task, there are only 4200 data sets, and more importantly, the ratio between the two categories is 9:1 2) Different principles of the model. Compared with first reason, we think the second reason is more important. As we can see, all models use the same data set but other algorithms do not have this problem. In contrast, Albert and BI-LSTM still present good performance. The former were 96.68% (UNT) and 95.45%(INT) in recall, another were 40.75% (UNT) and 96.24%(INT) in recall. This shows that the impact of the data set on the model is limited.

In SVM, words are not taken into account for their context meaning but are used as independent wholes to update weights and bigotry, which means that SVM classifies according to the similarity matching of words and sentences which considered as a whole. As shown in Figure 5.5, in the SVM model, the data is discretionally distributed, and the new data will be divided into two categories by the hyperplane determined according to the offensiveness of words learning from training set. This ignores the understanding of context and results in large errors in hyperplanes when offensive data are seriously lacking.

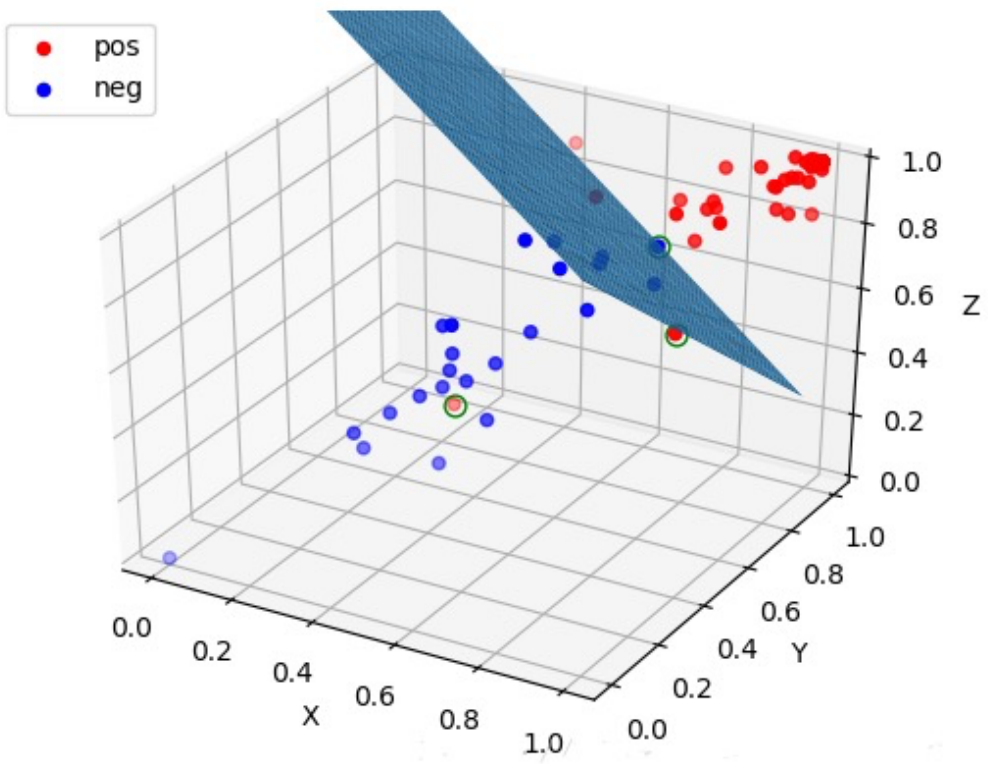

Figure 5.5: The principle of SVM

By contrast, Both BI-LSTM and Albert take context into account, but their approaches are different. As shown in Figure 2.8, BI-LSTM takes into account the sequence. When processing the data of the current moment, bi-LSTM consider the data of the previous moment, which increases the understanding of the following part. Bidirectional means that the LSTM will study one more time from the rear to the front, allowing the above relationship to be considered. However, LSTM has its limitations. As an improvement of RNN, the gradient explosion will still occur if the learning time sequence is too long.

The pre-training model ALbert is based on the attention mechanism, which not only focuses on the relationship between the words in this sentence, but also takes the sentences around it into consideration. Figure 5.6 shows the ALbert Atteition mechanism. It can be seen that the word 'the' pays attention not only to its own sentence, but also to the same words in other sentences and their relationships. Moreover, after a lot of text training, the pre-training model has mastered more patterns and corpus. This guarantees performance in downstream tasks.

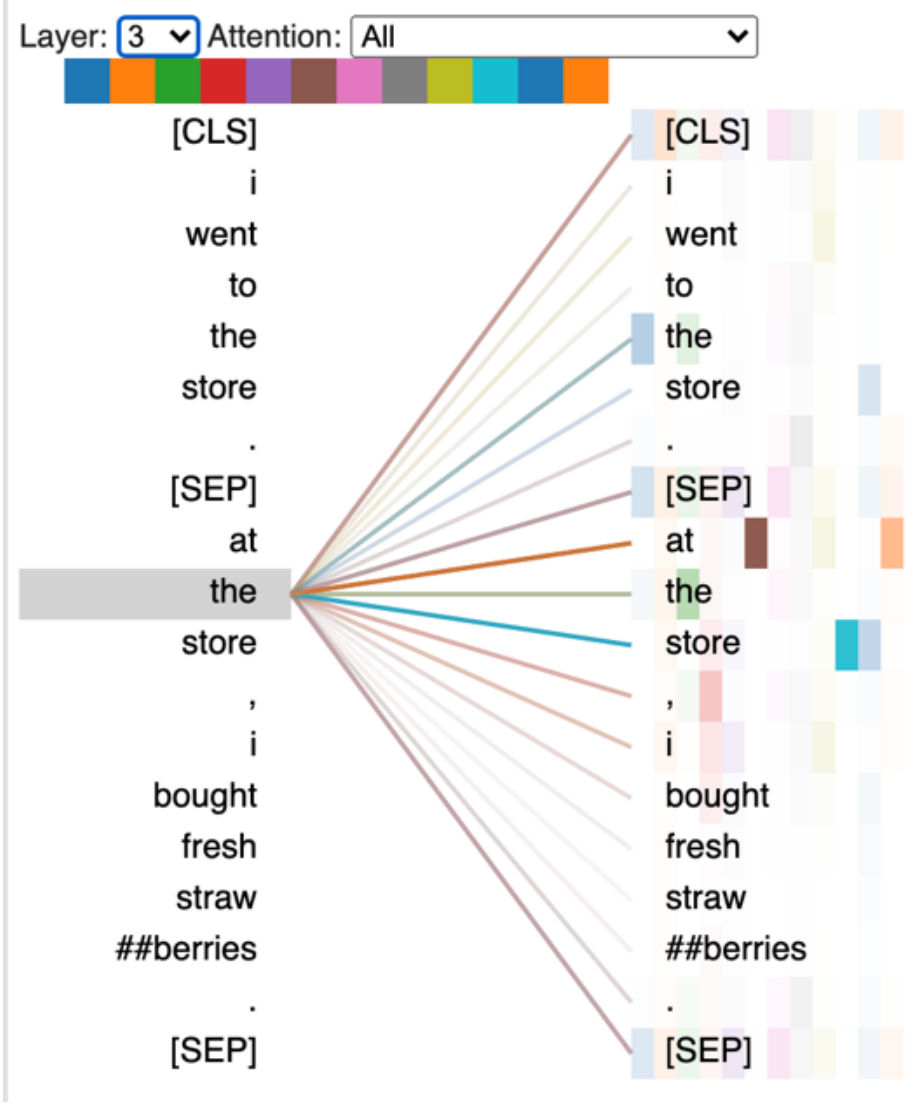

Figure 5.6: The principle of AlBert

ALbert's poor performance in subtask C was due to the lack of training time and data. The accuracy of SVM is similar to that of BI-LSTM, and its MCC score is 0.3369, which performs very well in eight models, which proves that SVM can also be used for multi-classification tasks.

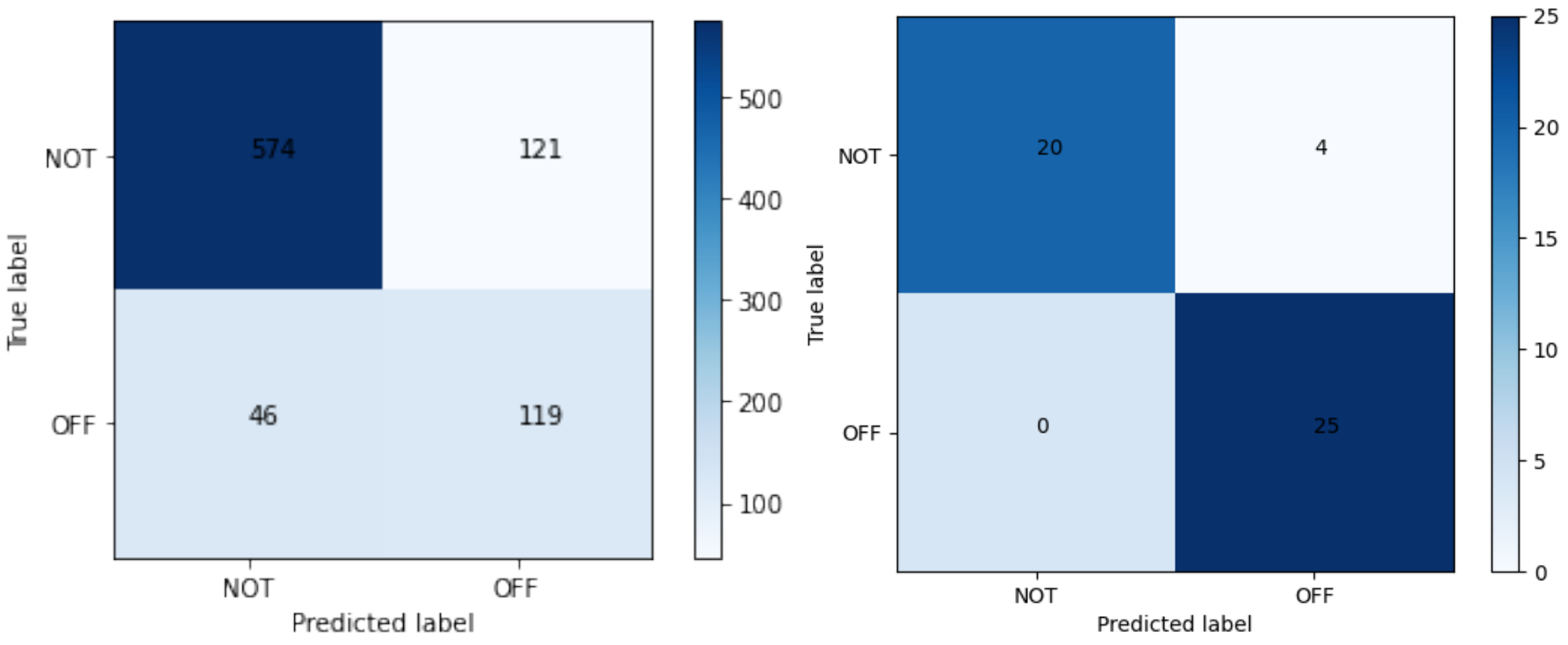

Figure 5.7: The confusion matrix of BI-LSTM and ALbert in subtask A

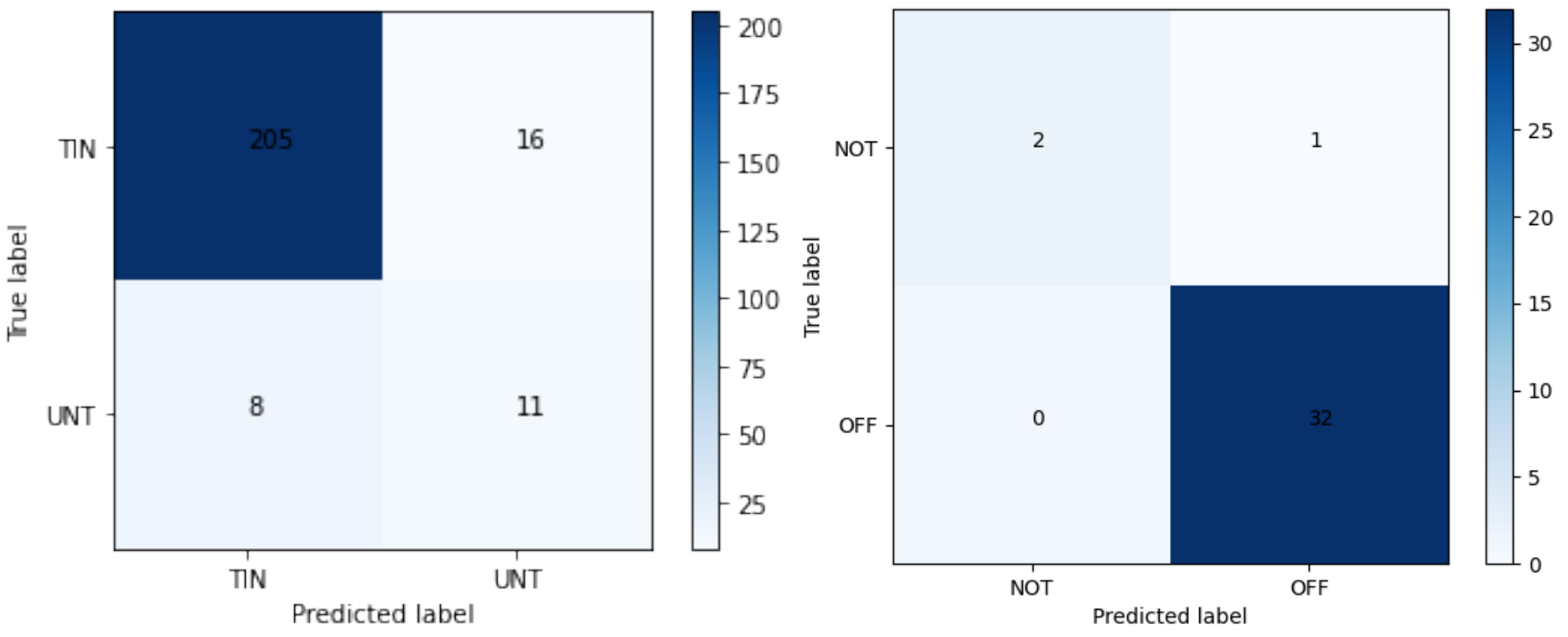

Figure 5.8: The confusion matrix of BI-LSTM and ALbert in subtask B

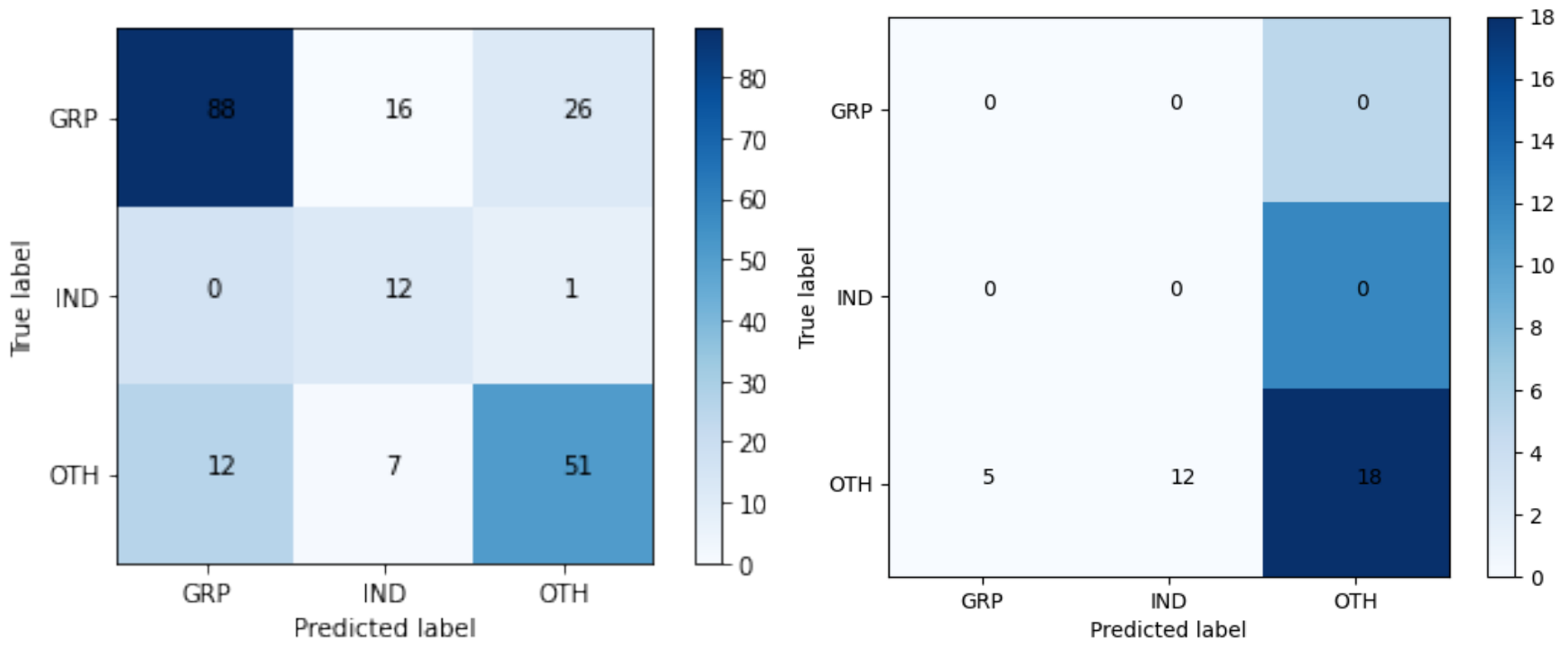

Figure 5.9: The confusion matrix of BI-LSTM and ALbert in subtask C

As the two models with the best performance, we draw the obturation matrix respectively to compare their performance. In Figure 5.7 and Figure 5.8, it is obvious that There are fewer errors in Albert reault. However, in multi-classification task C, BI-LSTM performed better, and ALbert even failed to classify two classes correctly. The reason for this phenomenon may be that we have not trained enough in the design of the model, and the randomly assigned training set may lead to a further reduction in the small number of the two types of samples.

# Chapter 6
# Conclusion

## 6.1 Overview

This report presents a variety of conceptual models for using Twitter data for offensive language recognition. This chapter summarizes the process and results of the project and how these results support or contradict the purpose and objectives of the project identified in Chapter 1. A key part of the project is goal setting. This chapter mainly describes the purpose of the project, whether the goals and objectives have been achieved. In addition, discuss the results of the project and check whether they have answered the research questions. Moreover, I also clarify and explain any limitations that prevent or contradict the expected assumptions. In the future work section, I will give some of these Suggestions that will help solve some of the problems related to offensive language. Another aspect of this chapter's explanation is to extract from the background the key facts involved in the research process and management of the project. Finally, I reflected on myself and what I had learned from the study.

## 6.2 Achievements and Limitations

1.In this project, I successfully identified the offensive language in the posts, completed three subtasks, and compared the performance of different algorithms through MCC, F1-Macro, confusion matrix and other methods. The code completed in this project includes data preprocessing, model building and model evaluation that can be directly used in future studies. The code of this project has good universality and high completion degree, and various data sets related to social media can be used in the future.

2.In this project, I implemented a comprehensive processing of the data set. Compared with the deletion of emoji in many studies, I translated it into words to express its content. Through a number of pretreatments, I have completed a clean data set that has been retained as far as possible and converted the preprocessed data into vectors by way of Word embedding. This means providing a data set that can be used directly for future research, saving the cost and time of preprocessing.

3. I describe the overall performance of the various algorithms on each task and the specific performance on each class. Subtask A ranks 30 in Team Rank, subtask B ranks 23, and subtask C ranks 32.

4. The pre-training model Albert training is too time-consuming, and therefore fails to get the best parameters during fine-tuning. In the future research, the time arrangement should be planned in advance to leave enough time for fine-tuning the model.

5. The lack of academic writing experience led me to revise my paper several times. In the future, it is necessary to increase the accumulation of research papers and improve the ability of academic writing.

### 6.3 Future Work

In future work, I recommend expanding the number of offensive posts in the data set to improve the training model as much as possible. At the same time, collect more targeted data. However, in order to do this, you need to understand and follow the legal provisions of Twitter, Facebook and other social medias, drink privacy policies and carefully consider the ethical implications. In addition, future research includes data preprocessing. Emoji is no longer a simple symbol, it has been given more than its own meaning. Translating emoji into its deeper meaning can be used to effectively improve the performance of data sets. Moverover, the improvement of the balance of data sets is also a research direction, and most of the current data sets have the problem of unbalanced data proportions. Generating or subtracting some qualified posts through certain means may be a research direction.

The optimization of existing algorithms is also a research direction. Classical classification algorithms such as SVM still have potential. By fine tuning and recording performance, good results can be obtained for specific tasks. Since 2012, deep learning models have been getting better and better at this task, and other deep learning models can be considered for problem solving. In recent years, the pre-training model has shown amazing performance in language classification tasks, which can be further developed by training the deep neural network model on the data set. Improving the pre-training model to make it more suitable for text classification can be a future research direction.

### 6.4 Self-reflection

This project has honed my skills in machine learning modeling,, text analysis, data mining and programming. In addition, I have managed to improve my research skills in effectively, quickly and correctly reading academic papers and journal articles by limiting my search to areas that are relevant to the topic I am working on.

While it is very exciting to complete this project, several challenges must be overcome in order to be successful. The main challenges in the early stages of this project were data preprocessing. Finding high-quality data has always been a laborious task in the field of deep learning. There is a lot of noise in the posts obtained on social media. Through investigating a large number of papers and websites, I found 11 methods of pre-processing and successfully completed high-quality data pre-processing task. One of the most enjoyable aspects of method is that instead of deleting emoji, as most researchers have done, it translates into a corresponding meaning. In addition, balancing model effectiveness and project completion is an important challenge. This involves the design and experiment of eight algorithms. Bert's training took longer than expected, and focusing only on it would make other models difficult to complete. Even for other models, many experiments are needed to select the optimal parameters.

I found myself very interested in the role of emoji in social media and the idea behind the model that was built in this project. If I have the opportunity to entry a PhD, I would like to further my study in this field, so that I can contribute my own ideas to this field one day.

Overall, in the process of this project, although experienced many difficulties and setbacks, but I overcame these, completed the goal on time, I also gained valuable experience.

## List of References

# Appendix A
# External Materials

### A.1 Project Code

The code for this project is in the following URL

Link: https://github.com/HarrywillDr/Identify-offensive-language-on-social-media

### A.2 Keras

The library used to build the model.

Link: https://keras.io/

### A.3 Tensorflow

This library is used to create Bi-LSTM and as the backend of Keras.

Link: https://www.tensorflow.org/

### A.4 Emoji

The code for pre-processing emoji.

Link:https://github.com/carpedm20/emoji

# Appendix B
# Ethical Issues Addressed

The data used in this project comes from the OLID dataset. The use of the data is strictly limited to academic research activities and does not derive any commercial benefit. The names of individuals and institutions in the data have been anonymised to avoid ethical issues in the data.